%% file: main.tex
\documentclass[runningheads]{llncs}

\usepackage{eccv}

\usepackage[dvipsnames]{xcolor}

\footnotetext[2]{$^*$ Equal contribution.\\$^\dag$ Corresponding author.}

\usepackage{eccvabbrv}

\usepackage{graphicx}
\usepackage{booktabs}
\usepackage{paralist}

\usepackage[accsupp]{axessibility}  

\usepackage{hyperref}
\hypersetup{
    colorlinks=true,
    urlcolor=magenta,
    citecolor=eccvblue,
    linkcolor=eccvblue,
}

\usepackage{orcidlink}

\usepackage{multirow}
\usepackage{ragged2e}
\usepackage{wrapfig}

\begin{document}

\title{Pyramid Diffusion for Fine 3D Large Scene Generation} 



\author{Yuheng Liu\inst{1,2}$^*$\orcidlink{0009-0001-3690-5659},
Xinke Li\inst{3}$^*$\orcidlink{0000-0002-9209-2154}, 
Xueting Li\inst{4}\orcidlink{0009-0009-2556-8667} ,
Lu Qi\inst{5}$^\dag$
\orcidlink{0000-0002-2684-0062},\\  
Chongshou Li\inst{1}\orcidlink{0000-0002-7595-0997},
Ming-Hsuan Yang\inst{5,6}\orcidlink{0000-0003-4848-2304}
}

\authorrunning{Y.~Liu et al.}

\institute{\mbox{%
\inst{1} Southwest Jiaotong University \quad
\inst{2} University of Leeds \quad
\inst{3} City University of Hong Kong}
\mbox{\inst{4} NVIDIA \quad
\inst{5} The University of California, Merced \quad
\inst{6} Yonsei University
}}

\maketitle

\begin{figure}
  \centering
  \includegraphics[width=0.95\linewidth]{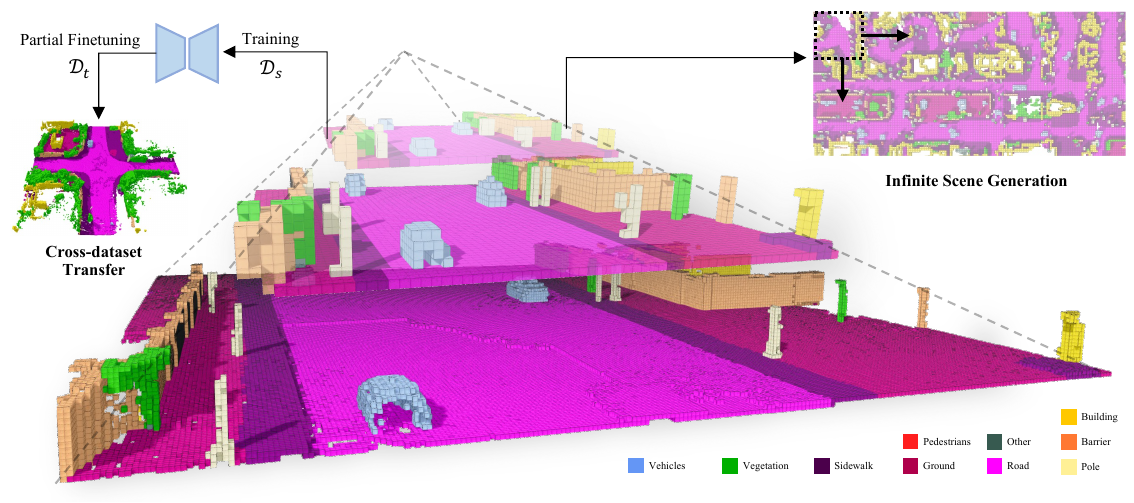}
\captionof{figure}{We present Pyramid Discrete Diffusion Model, a method that progresses from generating coarse- to fine-grained scenes, mirroring the top-down sequence of the pyramid structure shown. The model is extended for cross-dataset and infinite scene generation, with detailed scene intricacies illustrated on the flanking sides of the image. $\mathcal{D}_s$ and $\mathcal{D}_t$ refer to a source dataset and a target dataset, respectively.}
\label{fig:comp_coco}
\end{figure}

\input{sections/01.Abstract}

\input{sections/02.Introduction}

\input{sections/03.RelatedWork}

\input{sections/04.Approach}

\input{sections/05.Experiments}

\input{sections/06.Conclusion}

\clearpage  

\input{sections/07.Acknowledgements}

%
%
\bibliographystyle{splncs04}
\bibliography{main}

\input{sections/X_suppl}


\end{document}

%% file: sections/01.Abstract.tex
\begin{abstract}
Diffusion models have shown remarkable results in generating 2D images and small-scale 3D objects. However, their application to the synthesis of large-scale 3D scenes has been rarely explored. This is mainly due to the inherent complexity and bulky size of 3D scenery data, particularly outdoor scenes, and the limited availability of comprehensive real-world datasets, which makes training a stable scene diffusion model challenging. In this work, we explore how to effectively generate large-scale 3D scenes using the coarse-to-fine paradigm. We introduce a framework, the Pyramid Discrete Diffusion model (PDD), which employs scale-varied diffusion models to progressively generate high-quality outdoor scenes. Experimental results of PDD demonstrate our successful exploration in generating 3D scenes both unconditionally and conditionally. We further showcase the data compatibility of the PDD model, due to its multi-scale architecture: a PDD model trained on one dataset can be easily fine-tuned with another dataset. 
Code is available at \href{https://github.com/yuhengliu02/pyramid-discrete-diffusion}{https://github.com/yuhengliu02/pyramid-discrete-diffusion}.

\keywords{3D Scene Generation \and Diffusion Models \and Transfer Learning}

\end{abstract}

%% file: sections/02.Introduction.tex
\section{Introduction}
\label{sec:intro}
3D scene generation is the task of creating digital representations that mimic the three-dimensional complexities of our real-world environment, allowing for a more nuanced understanding of the tangible surroundings. This technique plays an essential role in fundamental computer vision tasks such as autonomous driving \cite{tang2023multi,li2022deepfusion,wu2022trajectory}, virtual reality \cite{sra2016procedurally,moro2021generation,ogun2019effect}, and robotic manipulation \cite{huang2023voxposer,cong2021comprehensive,xu2020learning}. However, high-quality large 3D scenes are extremely challenging to synthesize due to their inherently bulky size, and the lack of large-scale 3D scene datasets~\cite{wilson2022motionsc}. 

Meanwhile, recent advances in diffusion models have shown impressive results in generating 2D images~\cite{saharia2022photorealistic, rombach2022high, ramesh2022hierarchical} or small-scale 3D objects~\cite{poole2022dreamfusion, liu2023one2345++}. Yet, it is not a trivial task to employ diffusion models in 3D scene generation. On one hand, state-of-the-art diffusion models leave substantial memory footprints and demand considerable training time, posing a particular challenge when generating 3D scenes with large scales and intricate details. On the other hand, diffusion models require a large amount of training data~\cite{moon2022fine, zheng2023fast}, while capturing large-scale 3D scenes is itself a challenging and ongoing research topic~\cite{chen2023sd, lin2023infinicity}. As a result, only a few attempts have been made to apply diffusion models directly to 3D outdoor scenes~\cite{lee2023diffusion}, which resulted in unstable generation and thus suboptimal performance.

To resolve these challenges, existing works focus on conditional generation and resort to additional signals such as Scene Graphs~\cite{tang2023diffuscene} or 2D maps~\cite{mascaro2021diffuser} for guidance. Nonetheless, such conditional guidance is not always accessible, thereby restricting the generalizability of these approaches. Inspired by the coarse-to-fine philosophy widely used in image super-resolution \cite{sohldickstein2015deep, ho2020denoising, nichol2021improved}, we introduce the Pyramid Discrete Diffusion model (PDD), a framework that progressively generates large 3D scenes without relying on additional guidance.

We begin by generating small-scale 3D scenes and progressively increase the scale. At each scale level, we learn a separate diffusion model. This model takes the generated scene from the previous scale as a condition (except for the first diffusion model which takes noise as input) and synthesizes a 3D scene of a larger scale. Intuitively, this multi-scale generation process breaks down a challenging unconditional generation task (\ie, high-quality 3D scene generation) into several more manageable conditional generation tasks. This separation allows each diffusion model to specialize in generating either coarse structure (smaller scale) or intricate details (larger scale). Moreover, at the highest scale, we employ a technique known as scene subdivision, which involves dividing a large scene into multiple smaller segments that are synthesized using a shared diffusion model. This approach mitigates the issues of oversized models caused by the bulky size of the 3D scenes. Additionally, a noteworthy outcome of our multi-scale design is its capacity to facilitate cross-data transfer applications, which substantially reduces training resources. Lastly, we further propose a natural extension of our PDD framework with scene subdivision for infinite 3D scene generation, thereby demonstrating the scalability of the proposed method.

The main contributions of this work are as follows: 
\begin{itemize}
    \item We practically implement a coarse-to-fine strategy for 3D outdoor scene generation via designing a novel pyramid diffusion model.
    \item We conduct extensive experiments on our pyramid diffusion, demonstrating its generation of higher quality 3D scenes with comparable computational resources of existing approaches. In addition, we introduce new metrics to evaluate the quality of 3D scene generation from various perspectives. 
    \item Our proposed method showcases broader applications, enabling the generation of scenes from synthetic datasets to real-world data. Furthermore, our approach can be extended to facilitate the creation of infinite scenes.
\end{itemize}

%% file: sections/03.RelatedWork.tex
\section{Related Work}

\noindent\textbf{Diffusion Models for 2D Images.} 
Recent advancements in the generative model have seen the diffusion models \cite{sohldickstein2015deep, ho2020denoising, nichol2021improved} rise to prominence, especially in applications in 2D image creation \cite{dhariwal2021diffusion, rombach2022high, ramesh2022hierarchical}. 
In order to generate high-fidelity images via diffusion models, a multi-stage diffusion process is proposed and employed as per \cite{ho2022cascaded, saharia2022photorealistic, ho2022imagen}. This process starts with the generation of a coarse-resolution image using an initial diffusion model. Subsequently, a second diffusion model takes this initial output as input, refining it into a finer-resolution image. These cascaded diffusions can be iteratively applied to achieve the desired image resolution. 
We note that the generation of fine-grained 3D data presents more challenges than 2D due to the addition of an extra dimension. Consequently, our work is motivated by the aforementioned multistage 2D approaches to explore their applicability in 3D contexts. Furthermore, we aim to leverage the advantages of this structure to address the scarcity of datasets in 3D scenes.

\noindent\textbf{Diffusion Models for 3D Generation.} As a sparse and memory-efficient representation, 3D point clouds has been widely used in various computer vision applications such as digital human~\cite{Zheng2023pointavatar,su2023iccv,POP:ICCV:2021}, autonomous driving~\cite{li2020deep}, and 3D scene reconstruction~\cite{lan2019robust}. Point clouds generation aims to synthesize a 3D point clouds from a random noise~\cite{cheng2021learning,cheng2022autoregressive}, or scanned lidar points~\cite{lee2023diffusion}. Though the memory efficiency of point clouds is a valuable property, it poses high challenges in the task of point cloud generation. Existing works largely focus on using Generative Adversarial Networks (GANs), Variational Autoencoders (VAEs), or Vector Quantized Variational Autoencoders (VQ-VAEs) as the backbone for this task~\cite{cheng2022autoregressive,anvekar2022vg,cheng2021learning}. However, these models have limited capacity for high-fidelity generation and are notoriously known for unstable training. As an alternative to the generative models discussed above, diffusion models have revolutionized the computer vision community with their impressive performance in 2D image generation~\cite{saharia2022photorealistic, rombach2022high,ramesh2022hierarchical}. Yet, applying diffusion models for 3D point cloud generation has not been thoroughly explored hitherto. Point-Voxel Diffusion~\cite{Zhou_2021_ICCV} proposes to generate a raw point cloud through the diffusion process while LION~\cite{zeng2022lion} and DPM~\cite{luo2021diffusion} use the latent representation of a point cloud during the denoising process. However, all these methods focus on object-level point clouds and cannot be naively extended to scene-level point clouds. Most relevant to our work is~\cite{lee2023diffusion}, where a diffusion model is trained on a scene-level point cloud dataset for the synthesis task. However, due to the capacity limitation of diffusion models, generating a scene-level point cloud with a single diffusion model leads to unsatisfying results, such as undesired wholes or the lack of fine-grained objects. In this work, we propose a pyramid discrete diffusion model that reduces the difficulty at each pyramid level, thus producing scene point clouds with more realistic and fine-grained details.

\noindent\textbf{3D Large-scale Scene Generation.} Generating large-scale 3D scenes is an important but highly challenging task. A generative model on 3D scenes potentially provides infinite training data for tasks such as scene segmentation, autonomous driving, etc. Existing works~\cite{chen2023sd,li2022infinitenature,lin2023infinicity,xie2023citydreamer} simplify this task by first generating 2D scenes and then ``lifting'' them to 3D. Though such design is efficient for city scenes populated with regular geometries (e.g., buildings), it does not generalize easily to scenes with more fine-grained objects (e.g., pedestrians, cars, trees, etc.) In this paper, we directly generate 3D outdoor scenes using diffusion models, which include abundant small objects with semantics.

%% file: sections/04.Approach.tex
\begin{figure*}[t]
  \centering
  \includegraphics[width = 0.95\textwidth]
  {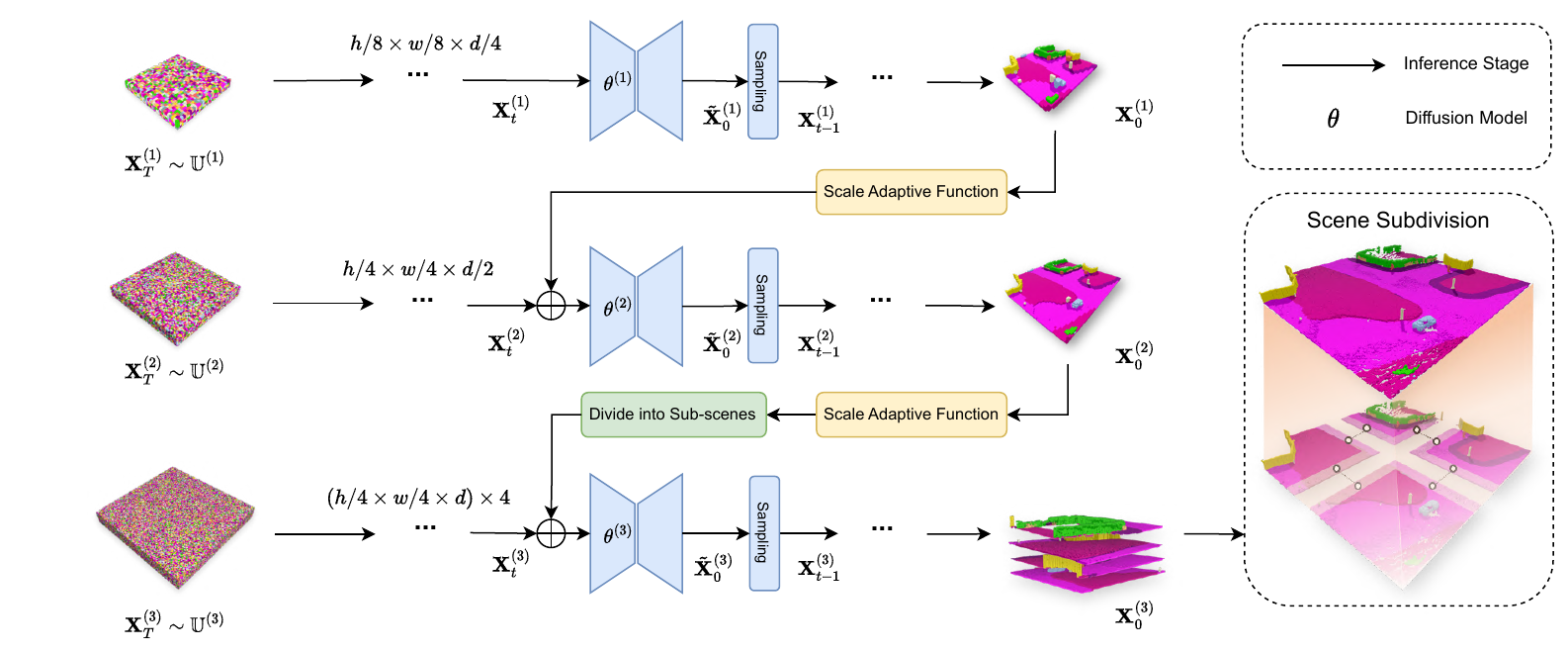}
    \caption{Framework of the proposed Pyramid Discrete Diffusion model. In our structure, there are three different scales. Scenes generated by a previous scale can serve as a condition for the current scale after processing through our scale adaptive function. Furthermore, for the final scale processing, the scene from the previous scale is subdivided into four sub-scenes. The final scene is reconstructed into a large scene using our Scene Subdivision module.}
    \label{fig:model-architecture}
\end{figure*}

\section{Approach}
The proposed  Pyramid Discrete Diffusion  (PDD) model comprises multi-scale models capable of step-by-step generation of high-quality 3D scenes from smaller scales. The PDD first extends the standard discrete diffusion for 3D data (Section~\ref{sec:3.2}) and then proposes a scene subdivision method to further reduce memory requirements (Section~\ref{sec:3.3}). Finally, we demonstrate two practical applications of PDD in specific scenarios (Section~\ref{sec:3.4}).

\subsection{Discrete Diffusion}\label{sec:3.1}

We focus on learning a data distribution based on  3D semantic scenes. Specifically, the semantic scene is represented in a one-hot format,  \textit{i.e.,} $\mathbf{X}\in \{0,1\}^{h \times w \times d \times c}$, where $h$, $w$, and $d$ indicate the dimensions of the scene, respectively, and $c$ denotes the size of the one-hot label.

Discrete diffusion \cite{austin2021structured} has been proposed to generate discrete data including semantic scenes. It involves applying the Markov transition matrix on discrete states for noise diffusion.  In the forward process, an original scene $\mathbf{X}_0$ is gradually corrupted into a $t$-step noised map $\mathbf{X}_t$ with $ t =1,\cdots,T$. Each forward step can be defined by a Markov uniform transition matrix $\mathbf{Q}_t$ as $\mathbf{X}_t=\mathbf{X}_{t-1} \mathbf{Q}_t$. Based on the Markov property, we can derive the $t$-step scene $\mathbf{X}_t$ straight from $\mathbf{X}_0$ with a cumulative transition matrix $\bar{\mathbf{Q}}_t=\mathbf{Q}_1 \mathbf{Q}_2 \cdots \mathbf{Q}_t$ :
\begin{equation}
q\left(\mathbf{X}_t \mid \mathbf{X}_0\right)=\operatorname{Cat}\left(\mathbf{X}_t ; \mathbf{P}=\mathbf{X}_0 \bar{\mathbf{Q}}_t\right), 
\end{equation}
where $\operatorname{Cat}(\mathbf{X}; \mathbf{P})$ is a multivariate categorical distribution over the one-hot semantic labels $\mathbf{X}$ with probabilities given
by $\mathbf{P}$. Finally, the semantic scene $\mathbf{X}_T$ at the last step $T$ is supposed to be in the form of a uniform discrete noise.
In the reverse process, a learnable model parametrized by $\theta$ is used to predict denoised semantic labels by $\tilde{p}_\theta\left(\tilde{\mathbf{X}}_{0} \mid \mathbf{X}_t\right)$. The reparametrization trick is applied subsequently to  get the reverse process $p_\theta\left(\mathbf{X}_{t-1} \mid \mathbf{X}_t\right)$ :
\begin{equation}
p_\theta\left(\mathbf{X}_{t-1} \mid \mathbf{X}_t\right) = \mathbb{E}_{\tilde{p}_\theta\left(\tilde{\mathbf{X}}_0 \mid \mathbf{X}_t\right)} q\left(\mathbf{X}_{t-1} \mid \mathbf{X}_t, \tilde{\mathbf{X}}_0\right).
\end{equation}
A loss consisting of the two KL divergences is proposed to learn better reconstruction ability for the model, given by
\begin{align}
\mathcal{L_{\theta}}&=d_{\text{KL}}\left(q\left(\mathbf{X}_{t-1} \mid \mathbf{X}_t, \mathbf{X}_0\right) \| p_\theta\left(\mathbf{X}_{t-1} \mid \mathbf{X}_t\right)\right)  \\ \notag
& + \lambda d_{\text{KL}}\left(q\left(\mathbf{X}_0\right) \| \tilde{p}_\theta\left(\tilde{\mathbf{X}}_0 \mid \mathbf{X}_t\right)\right),
\end{align}
where $\lambda$ is an auxiliary loss weight and $d_{\text{KL}}$ stands for KL divergence. In the following, we focus on extending the discrete diffusion into the proposed PDD.

\subsection{Pyramid Discrete Diffusion}\label{sec:3.2}
We propose PDD that operates various diffusion processes across multiple scales (or resolutions),  as depicted in Figure \ref{fig:model-architecture}. Given a 3D scene data $\mathbf{Z}\in \{0,1\}^{h\times w \times d \times c}$, we define a 3D pyramid including  different scales of  $\mathbf{Z}$, \textit{i.e.,} $\{\mathbf{Z}^{(1)}, \cdots, \mathbf{Z}^{(l)},\cdots, \mathbf{Z}^{(L)}\}$, 
where a larger $l$ indicates a larger scene scale. Formally, let $h_l\times w_l \times d_l \times c$ denote the dimension of $\mathbf{Z}^{(l)}$,
 $h_{l+1}\geq h_{l}$, $w_{l+1}\geq w_{l}$ and $d_{l+1}\geq d_{l}$ are kept for $l=1,\cdots, L-1$. We note that such a pyramid can be obtained by applying different down-sample operators, such as pooling functions, on $\mathbf{Z}$. 
For each scale in the pyramid, we construct a conditional discrete diffusion model parameterized by $\theta_{l}$. The $l$-th model for $l\neq 1$ is given by: 
\begin{equation}
 \tilde{p}_{\theta_{l}} \left(\tilde{\mathbf{X}}_0^{(l)} \mid \mathbf{X}_t^{(l)}, \mathbf{Z}^{(l-1)} \right) = \tilde{p}_{\theta_{l}}\left(\tilde{\mathbf{X}}_0^{(l)} \mid \text{Concat}\left(\mathbf{X}_t^{(l)}, \phi^{(l)}(\mathbf{Z}^{(l-1)})\right) \right), 
\end{equation}
where  $\mathbf{X}_t^{(l)}$ and  $\mathbf{X}_0^{(l)}$ are with the same size of   $\mathbf{Z}^{(l)}$ , and $\phi^{(l)}$ is a Scale Adaptive Function (SAF) for upsamling $\mathbf{Z}^{(l-1)}$ into the size of  $\mathbf{Z}^{(l)}$.  As a case in point, SAF can be a tri-linear interpolation function depending on the data. Additionally, we maintain the first model $\tilde{p}_{\theta_{1}}$ as the original non-conditional model.

During the training process, PDD learns $L$ denoising models separately at varied scales of scene pyramids in the given dataset. Given that $\mathbf{Z}^{(l-1)}$  is essentially a lossy-compressed version of $\mathbf{Z}^{(l)}$ , the model training can be viewed as learning to restore the details of a coarse scene. In the inference process,  denoising model $p_{\theta_1}$ is performed initially according to Equation (2) and the rest of PDD models are executed in sequence from $l=2$ to $L$ via the sampling,
\begin{align}
 \mathbf{X}_{t-1}^{(l)} \sim p_{\theta_l}(\mathbf{X}_{t-1}^{(l)} \mid \mathbf{X}_{t}^{(l)}, \mathbf{X}_{0}^{(l-1)}),
\end{align}
 where $ \mathbf{X}_{0}^{(l-1)}$ is the denoised result of $\tilde{p}_{\theta_{l-1}}$.

Except for the high-quality generation, the proposed PDD bears two merits: 1) Diffusion models in PDD can be trained in parallel due to their independence, which allows for a flexible computation reallocation during training. 2) Due to its multi-stage generation process, PDD is fitting for restoring scenes of arbitrary coarse-grained scale by starting from the intermediate processes, thereby extending the method's versatility. 

\subsection{Scene Subdivision}\label{sec:3.3}
To overcome the memory constraint for generating large 3D scenes, we propose the scene subdivision method. We divide a 3D scene $\mathbf{Z}^{(l)}$  along $z$-axis into $I$ overlapped sub-components as $\{ \mathbf{Z}_{i}^{(l)} \}_{i=1}^{I}$. For the instance of four subscenes case, let $\mathbf{Z}_{i}^{(l)}\in \{0,1\}^{ (1+\delta_l)h_l\backslash 2 \times (1+\delta_l)w_l\backslash 2 \times d_l \times c }$ denote one subscene and $\delta_l$ denote the overlap ratio, the shared $l$-th diffusion model in PDD is trained to reconstruct $\mathbf{Z}_{i}^{(l)}$ for $i=1,\cdots, 4$. 
Subsequently, sub-scenes are merged into a holistic one by a fusion algorithm, \textit{i.e.}, voting on the overlapped parts to ensure the continuity of the 3D scene.

In the training process, to ensure context-awareness of the entire scene during the generation of a sub-scene, we train the model by adding the overlapped regions with other sub-scenes as the condition.  In the inference process, the entire scene is generated in an autoregressive manner.  Apart from the first sub-scene generated without context, all other sub-scenes utilize the already generated overlapped region as a condition, \textit{i.e.},
\begin{equation}
 \mathbf{X}_{t-1, i}^{(l)} \sim p_{\theta} \left( \mathbf{X}_{ t-1, i}^{(l)} \mid \mathbf{X}_{t, i}^{(l)}, \mathbf{X}_{0, i}^{(l+1)} , \sum_{j\neq i}\Delta_{ij}\odot \mathbf{X}_{
 0, j}^{(l+1)}\right),
\end{equation}
where $j$ is the index of generated sub-scenes before $i$-th scene, and $\Delta_{ij}$ is a binary mask between $\mathbf{X}_{0, i}^{(l+1)}$ and $\mathbf{X}_{0, j}^{(l+1)}$ representing the overlapped region on $\mathbf{X}_{0, j}^{(l+1)}$  with 1 and the separate region with 0. Scene Subdivision module can reduce the model parameters, as diffusion model could be shared by four sub-scenes. In practice, we only implement the scene subdivision method on the largest scale which demands the largest memory.

\subsection{Applications}\label{sec:3.4}
Beyond its primary function as a generative model, we introduce two novel applications for PDD.
First, \textbf{cross-dataset transfer} aims at adapting a model trained on a source dataset to a target dataset~\cite{zhuang2020comprehensive}. Due to the flexibility of input scale, PDD can achieve this by retraining or fine-tuning the smaller-scale models in the new dataset while keeping the larger-scale models. The strategy leveraging PDD  improves the efficiency of transferring 3D scene generation models between distinct datasets. 
Second,\textbf{ infinite scene generation} is of great interest in fields such as autonomous driving~\cite{geiger2012we} and urban modeling~\cite{li2020campus3d} which require a huge scale of 3D scenes. PDD can extend its scene subdivision technique. By using the edge of a previously generated scene as a condition as in Equation (6), it can iteratively create larger scenes, potentially without size limitations.

%% file: sections/05.Experiments.tex
\section{Experimental Results}
\subsection{Evaluation Protocols}
Since the metrics used in  2D generation such as FID~\cite{heusel2017gans} are not directly applicable in the 3D, we introduce and implement three metrics to assess the quality of the generated 3D scenes. We note that more implementation details can be found in the supplementary material. 

\begin{table*}[t]
\caption{Comparison of various diffusion models on 3D semantic scene generation of CarlaSC. DiscreteDiff \cite{austin2021structured}, LatentDiff \cite{lee2023diffusion}, and P-DiscreteDiff refer to the original discrete diffusion, latent discrete diffusion, and our approach, respectively. Conditioned models work based on the context of unlabeled point clouds or the coarse version of the ground truth scene. A higher \textit{Segmentation Metric} value is better, indicating semantic consistency. A lower \textit{Feature-based Metric} value is preferable, representing closer proximity to the original dataset. The brackets with \textit{V} represent voxel-based network and \textit{P} represent point-based network.}\label{comparison-models-generation}
\footnotesize
\centering
\resizebox{0.99\textwidth}{!}{
\begin{tabular}{lcccccccc}
\toprule
\multirow{2}{*}{\textbf{Method}}        & \multirow{2}{*}{\textbf{Model}}            & \multirow{2}{*}{\textbf{Condition}} & \multicolumn{4}{c}{\textbf{Segmentation Metric}}                                                                              & \multicolumn{2}{c}{\textbf{Feature-based Metric}}          \\
                               &                                   &                            & \multicolumn{1}{c}{mIoU (V)} & \multicolumn{1}{c}{MA (V)} & \multicolumn{1}{c}{mIoU (P)} & \multicolumn{1}{c}{MA (P)} & \multicolumn{1}{c}{F3D ($\downarrow$)} & \multicolumn{1}{c}{MMD ($\downarrow$)} \\ \midrule \midrule
Ground Truth                   & -                                 & -                          &                              52.19&                            72.40&                              32.90&                            47.68&                         0.246&                         0.108\\ \midrule
\multirow{3}{*}{Unconditioned} & DiscreteDiff \cite{austin2021structured}                & -                          &                              40.05&                            63.65&                              25.54&                            38.71&                         1.361&                         0.599\\
                               & LatentDiff \cite{lee2023diffusion}                 & -                          &                              38.01&                            62.39&                              26.69&                            45.87&                         0.331&                         0.211\\
                               & P-DiscreteDiff (Ours) & -                          &                              \textbf{68.02}&                            \textbf{85.66}&                              \textbf{33.89}&                            \textbf{52.12}&                         \textbf{0.315}&                         \textbf{0.200}\\ \midrule
\multirow{3}{*}{Conditioned}& DiscreteDiff \cite{austin2021structured}              & Point cloud                &                              38.55&                            59.97&                              28.41&                            44.06&                         0.357&                         \textbf{0.261}\\
 & DiscreteDiff \cite{austin2021structured} & Coarse scene ($s_1$)  & 52.52& 77.23& 27.93& 43.13& 0.359&0.284\\
                               & P-DiscreteDiff (Ours) & Coarse scene ($s_1$)                 &                              \textbf{55.75}&                            \textbf{78.70}&                              \textbf{29.78}&                            \textbf{46.61}&                         \textbf{0.342}&                         0.274\\\bottomrule
\end{tabular}
}
\end{table*}

\textnormal{Semantic Segmentation} results on the generated scenes are used to evaluate the effectiveness of models in creating semantically coherent scenes. Specifically, two architectures, the voxel-based SparseUNet~\cite{graham20183d} and point-based PointNet++~\cite{qi2017pointnet++}, are implemented to perform the segmentation tasks. We report the mean Intersection over Union (mIoU) and Mean Accuracy (MAs) for evaluation.

\textnormal{F3D} is a  3D adaption of the 2D Fréchet Inception Distance (FID)~\cite{heusel2017gans}, which is based on a pre-trained autoencoder with an 3D CNN architecture. We calculate and report the Fréchet distance (by $10^{-3}$ ratio) between the generated scenes and real scenes in the feature domain.

 \textnormal{Maximum Mean Discrepancy} (MMD) is a statistical measure to quantify the disparity between the distributions of generated and real scenes. Similar to our F3D approach, we extract features via the same pre-trained autoencoder and present the MMD between 3D scenes.

\subsection{Experiment Settings}
\noindent \textbf{Datasets.} 
We use CarlaSC~\cite{wilson2022motionsc} and SemanticKITTI~\cite{behley2019semantickitti} for experiments. 
Specifically, we conduct our main experiments as well as ablation studies on the synthesis dataset CarlaSC due to its large data volume and diverse semantic objects. Our primary model is trained on the training set of CarlaSC with 10 categories and 32,400 scans. 
SemanticKITTI, which is a real-world collected dataset, is used for our cross-dataset transfer experiment. 
 Both datasets are adjusted to ensure consistency in semantic categories, with further details in the supplementary material. 

\noindent \textbf{Model Architecture.} The primary proposed PDD is performed on three scales of a 3D scene pyramid, \ie, $s_1$, $s_2$ and  $s_4$ in Table~\ref{tab:scales}. We implement 3D-UNets~\cite{cciccek20163d} for three diffusion models in PDD based on the scales.  Notably, the model applied on $s_4$ scale is with the input/output size of $s'_3$ due to the use of scene subdivision, while such a size of other models follows the working scale size. In the ablation study, we also introduce the scale $s_3$ in the experiment. Additionally, we implement two baseline methods merely on scale $s_4$ which are the original discrete diffusion~\cite{austin2021structured} and the latent diffusion model with VQ-VAE decoder~\cite{lee2023diffusion}.

\noindent \textbf{Training Setting.} 
We train each PDD model using the same training setting except for the batch size. Specifically, we set the learning rate of $10^{-3}$ for the AdamW optimizer~\cite{loshchilov2017decoupled}, and the time step $T = 100$ for the diffusion process, and 800 for the max epoch. The batch sizes are set to 128, 32, and 16 for the models working on $s_1$, $s_2$ and $s_4$ scales. However, for the baseline method based on the $s_4$ scale, we use the batch size of 8 due to memory constraints. We note that all diffusion models are trained on four NVIDIA A100 GPUs. In addition, we apply the trilinear interpolation for the scene fusion algorithm and set the overlap ratio in scene subdivision,  $\delta_l$ to 0.0625. 

\begin{figure}[tp]
\begin{minipage}{0.45\textwidth}
\centering
\captionof{table}{Different scales in the 3D scene pyramid.}\label{tab:scales}
\setlength{\tabcolsep}{5pt}
\begin{tabular}{cc}
\toprule
\textbf{Scale Rep.} & \textbf{3D Scene Size} \\ \midrule \midrule
$s_1$ & $32 \times 32 \times 4$ \\
$s_2$ & $64 \times 64 \times 8$ \\
$s_3$ & $128 \times 128 \times 8$ \\
$s'_3$ & $136 \times 136 \times 16$ \\
$s_4$ & $256 \times 256 \times 16$ \\ \bottomrule
\end{tabular}
\end{minipage}%
\hfill
\begin{minipage}{0.48\textwidth}
    \centering
    \captionof{table}{Average SSIM between each generated scene and the closest scene in the training set. We generate 1$k$ scenes and calculat the average.}
    \label{tab:ssim}
    \resizebox{0.85\textwidth}{!}{%
        \setlength{\tabcolsep}{9pt}
        \begin{tabular}{llc}
        \toprule
        \textbf{Dataset}                    & \textbf{Method}    & \textbf{SSIM} \\ \midrule \midrule
        \multirow{2}{*}{CarlaSC}   & Uncon-Gen & 0.72 \\
                           & Val Set   & 0.65 \\ \midrule
        Sem-KITTI & Uncon-Gen & 0.74 \\
        (\textit{fine-tuned})
                           & Val Set   & 0.67 \\ \bottomrule
        \end{tabular}
    }
\end{minipage}
\end{figure}

\begin{table}[t]
\captionof{table}{Comparison of different diffusion pyramids on 3D semantic scene generation.}\label{multi-stage}
\centering
\resizebox{0.9\textwidth}{!}{
\setlength{\tabcolsep}{5pt}
\begin{tabular}{ccccccc}
\toprule
\textbf{Pyramid}  & \textbf{Conditioned} & \textbf{mIoU(V)} & \textbf{mIoU(P)} & \textbf{F3D($\downarrow$)} & \textbf{MMD($\downarrow$)} \\
\midrule \midrule
\(s_4\) & \(\times\) & 40.0 & 25.5 & 1.36 & 0.60 \\
\(s_1 \rightarrow s_4\) & \(\times\) & 67.0 & 32.1 & 0.32 & 0.24 \\
\(s_1 \rightarrow s_2 \rightarrow s_4\) & \(\times\) & \textbf{68.0} & \textbf{33.9} & \textbf{0.32} & \textbf{0.20} \\
\(s_1 \rightarrow s_2 \rightarrow s_3 \rightarrow s_4\) & \(\times\) & \textbf{68.0} & 33.4 & \textbf{0.32} & 0.23 \\ \midrule
\(s_1 \rightarrow s_4\) & \(\checkmark\) & 52.5 & 27.9 & 0.36 & 0.28 \\
\(s_1 \rightarrow s_2 \rightarrow s_4\) & \(\checkmark\) & 55.8 & \textbf{29.8} & \textbf{0.34} & \textbf{0.27} \\
\(s_1 \rightarrow s_2 \rightarrow s_3 \rightarrow s_4\) & \(\checkmark\) & \textbf{55.9} & 29.6 & \textbf{0.34} & 0.28 \\
\bottomrule
\end{tabular}}
\end{table}

\subsection{Main Results}

\begin{figure*}[t]
  \centering
  \includegraphics[width = 0.99\textwidth]
  {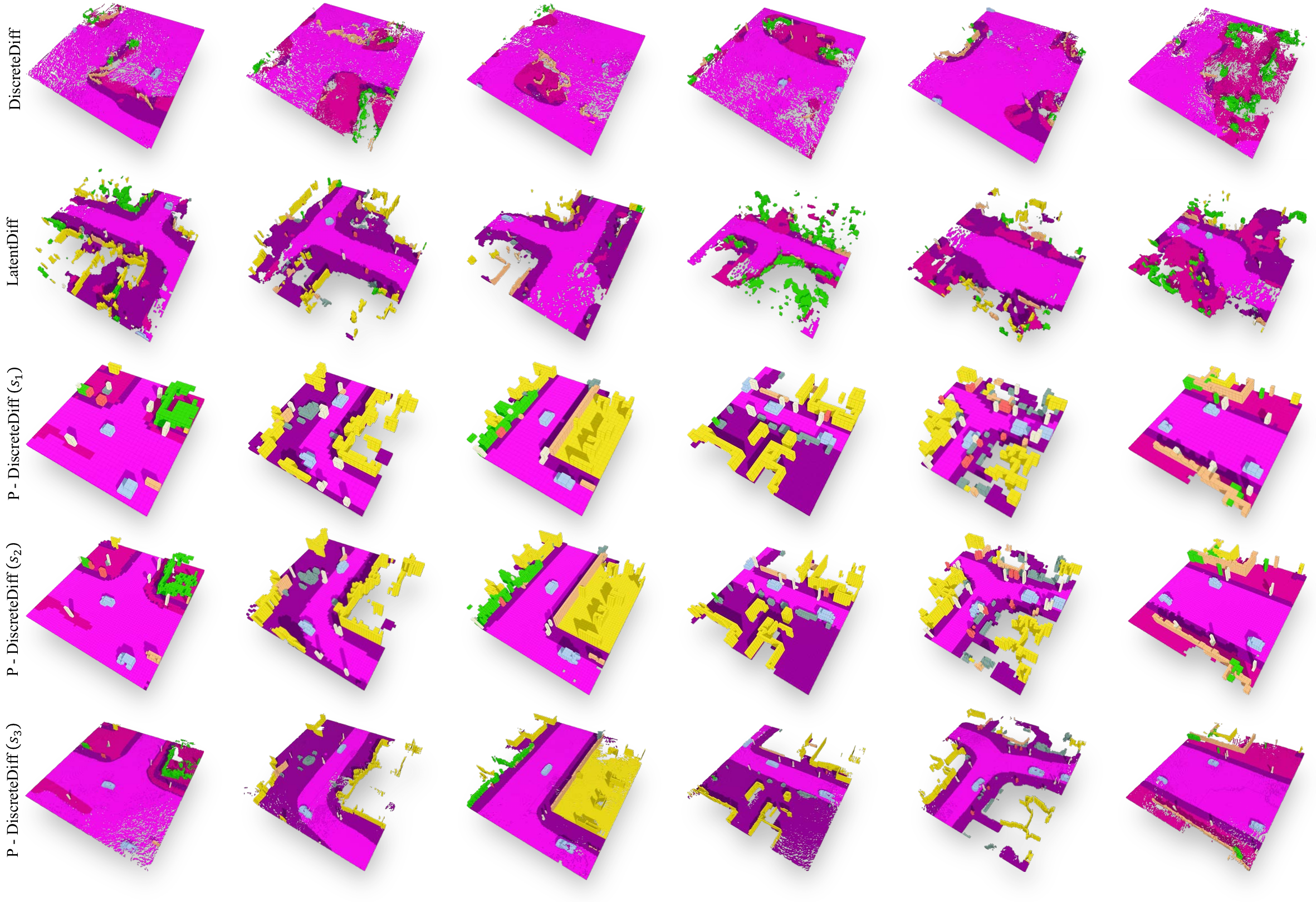}
    \caption{Visualization of unconditional generation results on CarlaSC. We compare with two baseline models -- DiscreteDiff \cite{austin2021structured} and LatentDiff \cite{lee2023diffusion} and show synthesis from our models with different scales. Our method produces more diverse scenes compared to the baseline models. Furthermore, with more levels, our model can synthesize scenes with more intricate details.}
    \label{fig:unconditional-generation}
\end{figure*}

\noindent\textbf{Generation Quality.} We compare our approach with two baselines, the original Discrete Diffusion \cite{austin2021structured} and the Latent Diffusion \cite{lee2023diffusion}. The result reported in Table~\ref{comparison-models-generation} demonstrates the notable performance of our method across all metrics in both unconditional and conditional settings in comparable computational resources with existing method. Our proposed method demonstrates a notable advantage in segmentation tasks, especially when it reaches around 70\% mIoU for SparseUNet, which reflects its ability to generate scenes with accurate semantic coherence.  We also provide visualizations of different model results in Figure \ref{fig:unconditional-generation}, where the proposed method demonstrates better performance in detail generation and scene diversity for random 3D scene generations.

\begin{figure*}[t]
  \centering
  \includegraphics[width = 0.99\linewidth]
  {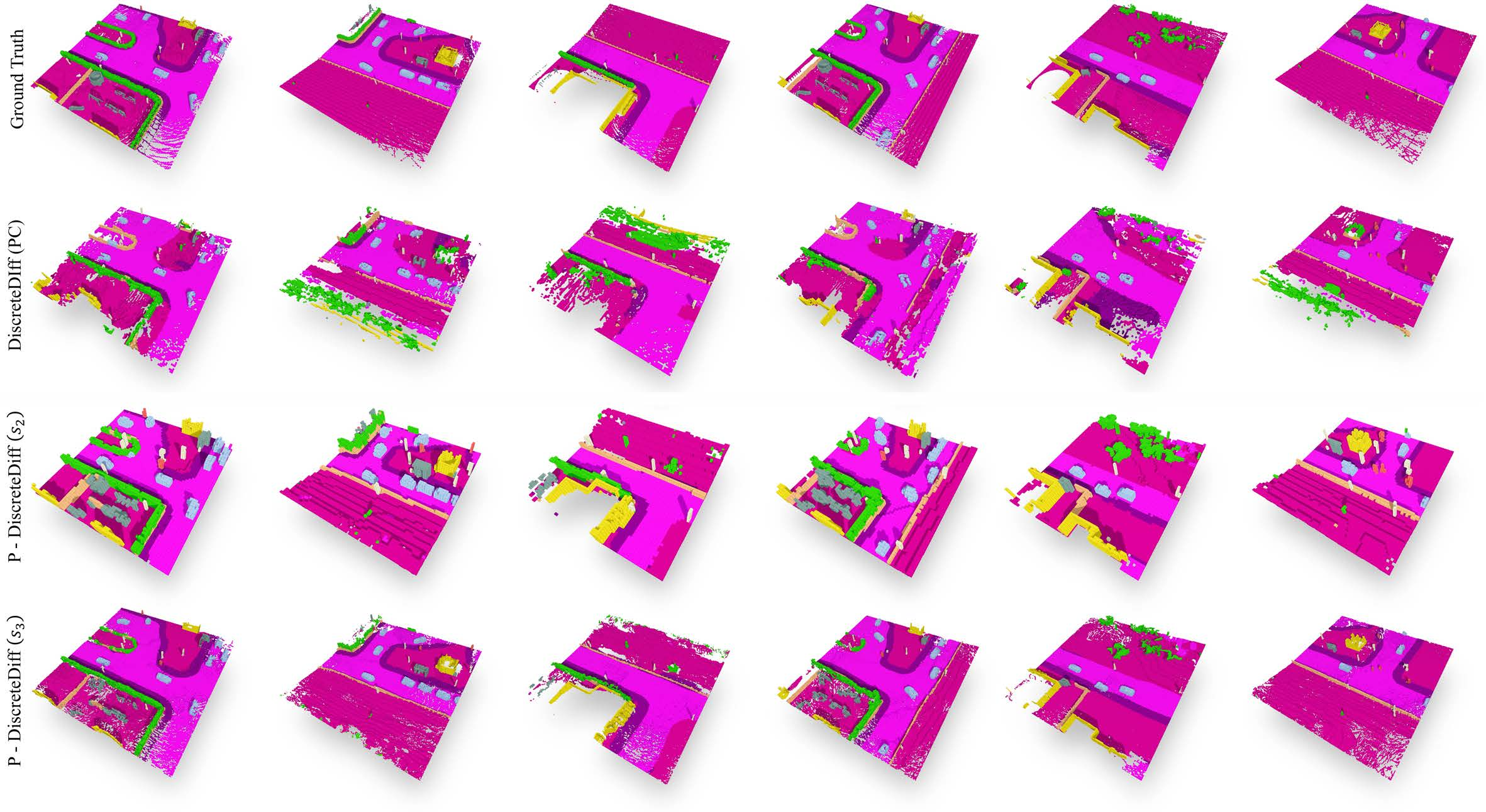}
    \caption{Visualization of conditional generation results on CarlaSC. \textit{PC} stands for point cloud condition.}
    \label{fig:conditional-generation}
\end{figure*}

Additionally, we conduct the comparison on conditioned 3D scene generation. We leverage the flexibility of input scale for our method and perform the generation by models in $s_2$ and $s_4$ scales conditioned on a coarse ground truth scene in $s_1$ scale.  We benchmark our method against the discrete diffusion conditioned on unlabeled point clouds and the same coarse scenes. Results in Table~\ref{comparison-models-generation} and Figure \ref{fig:conditional-generation} present the impressive results of our conditional generation comparison.  It is also observed that the point cloud-based model can achieve decent performance on F3D and MMD, which could be caused by 3D point conditions providing more structural information about the scene than the coarse scene. Despite the informative condition of the point cloud, our method can still outperform it across most metrics.

\begin{figure*}[t]
  \centering
  \includegraphics[width = 0.99\textwidth]
  {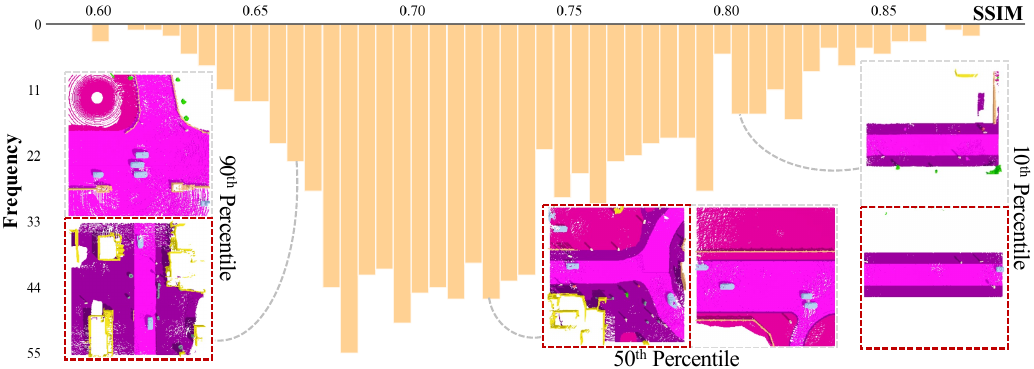}
    \caption{Data retrieval visualization. We generate 1$k$ scenes using PDD on CarlaSC dataset, and retrieve the most similar scene in training set for each scene using SSIM (SSIM=1 means identical), and plot SSIM distribution. Scenes at various percentiles are displayed (red box: generated scenes; grey box: scenes in training set), those with the highest 10\% similarity (i.e., 10$^{th}$ percentile of SSIM) are very similar to the training set, but still not completely identical.}
    \label{fig:rebuttal_vis2}
\end{figure*}

\noindent\textbf{None-overfitting Verification.} 
The MMD and F3D metrics (defined in Supp A) numerically illustrate the statistical feature distance between generated scenes and the training set. Our method achieves the lowest MMD and F3D among all baseline methods as shown in Table \ref{comparison-models-generation}. However, we argue that this \textit{does not indicate overfitting} to the dataset for the following reasons. First, our MMD and F3D are larger than those of the ground truth. Furthermore, we leverage structural similarity (SSIM) to search and show that a generated scene is different from its nearest neighbour in the dataset.
Specifically, we generate 1$k$ scenes and identify their closest matches in the training set using the SSIM metric. 
The average SSIM of these 1$k$ scenes are calculated and presented in Table \ref{tab:ssim}. 
Additionally, we apply the same methodology to the Validation Set to establish an oracle baseline. 
Table \ref{tab:ssim} shows that our generated scenes are comparable to the baseline, verifying that our method does not overfit the training set.
To further support this, we use distribution plots, as shown in Figure \ref{fig:rebuttal_vis2}, to validate the similarity between our generated scenes and the training set. 
We also display three pairs at different percentiles in this figure, showing that scenes with lower SSIM scores differ more from their nearest matches in the training set. 
This visual evidence reinforces that PDD effectively captures the distribution of the training set instead of merely memorizing it.

\subsection{Ablation Studies}

\noindent\textbf{Pyramid Diffusion.} Our experiments explore the impact of varying refinement scales on the quality of generated scenes. According to Table \ref{multi-stage}, both conditional and unconditional scene generation quality show incremental improvements with additional scales. Balancing training overhead and generation quality, a three-scale model with the scale of $s_4$ progression offers an optimal compromise between performance and computational cost. We find that as the number of scales increases, there is indeed a rise in performance, particularly notable upon the addition of the second scale. However, the progression from a three-scale pyramid to a four-scale pyramid turns out to be insignificant. Given the greater training overhead for a four-scale pyramid compared to a three-scale one, we choose the latter as our main structure.

\begin{figure}[t]
    \centering
    \includegraphics[width = 0.99\linewidth]
    {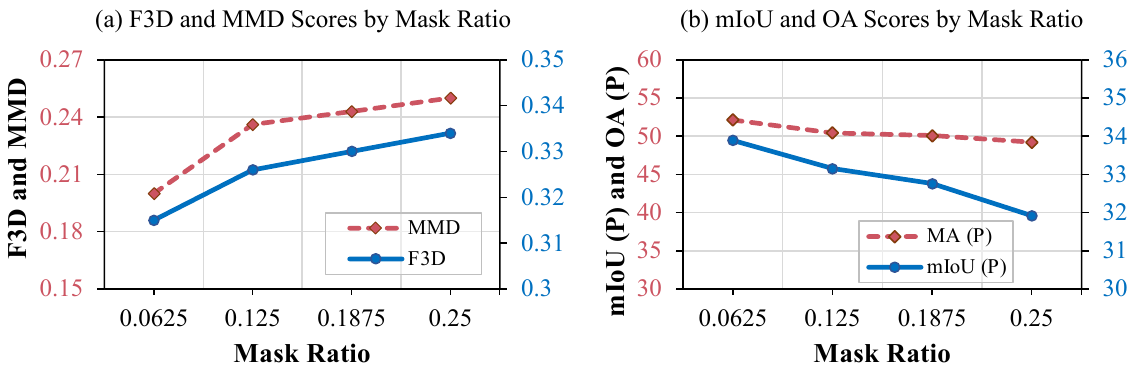}
    \caption{Effects of mask ratio on unconditional generation results.}
    \label{fig:line-chary} 
\end{figure}

\noindent\textbf{Scene Subdivision.} 
We explore the optimal mask ratio for scene subdivision and report on Figure \ref{fig:line-chary}, which shows an inverse correlation between the mask ratio and the effectiveness of F3D and MMD metrics; higher mask ratios may result in diminished outcomes due to increased overfitting, leading to reduced randomness in the generated results. The lowest mask ratio test, 0.0625, achieves the best results across all metrics with detailed retention. Thus, we set a mask ratio of 0.0625 as the standard for our scene subdivision module. Further analysis shows that higher overlap ratios in scene subdivision result in quality deterioration, mainly due to increased discontinuities when merging sub-scenes using scene fusion algorithm.

\noindent\textbf{Performance on Different Scales.}
To facilitate a comprehensive understanding of the progressive improvement achieved by our coarse-to-fine method, we evaluate the quality of scenes generated at different scales and present the findings in Table~\ref{different-stages}. Even at the smaller scale $s_1$, we observe high F3D and MMD scores, indicating its capacity in synthesizing scenes with both reasonable and diverse layouts. As we advance to larger scales (i.e., $s_2$ and $s_3$), the mIoU and MA scores consistently increase, demonstrating that our model focuses on learning intricate details in later stages. Meanwhile, F3D and MMD metrics show stability without significant decline, indicating a balanced enhancement in scene complexity and fidelity.

\begin{table}[t]
\caption{Generation results on CarlaSC in different scales on the diffusion pyramid without any conditions. All output scales are lifted to $s_4$ using the upsampling method.}
\label{different-stages}
\centering
\footnotesize
\setlength{\tabcolsep}{5pt}
\begin{tabular}{cccccc}
\toprule
 \textbf{Model No.} & \textbf{Scale} & \textbf{mIoU($\uparrow$)}& \textbf{MA($\uparrow$)}& \textbf{F3D($\downarrow$)}    & \textbf{MMD($\downarrow$)}    \\ \midrule \midrule
1        & $s_1$& 18.0 & 42.7& 0.29& 0.16\\
2        & $s_2$& 43.7 & 66.8& 0.29& 0.18\\
3        & $s_4$    & 68.0& 85.7& 0.32& 0.23\\ \bottomrule
\end{tabular}
\end{table}

\subsection{Applications}
\label{sec::application}
\begin{figure}[t]
    \centering
    \begin{minipage}[t]{0.48\textwidth}
        \centering
        \includegraphics[width=0.95\linewidth]{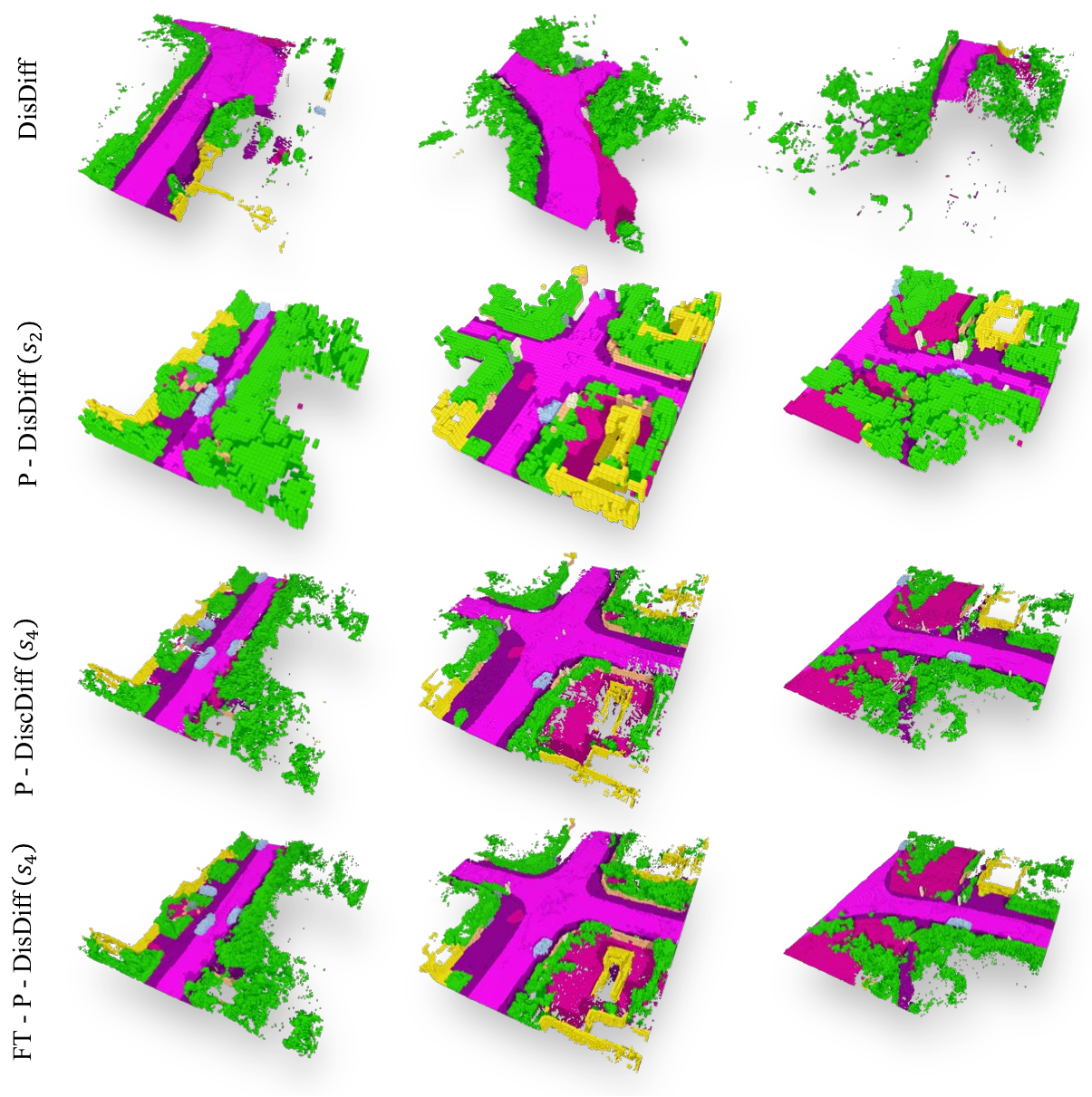}
        \caption{SemanticKITTI unconditional generation. \textit{FT} stands for finetuning pre-trained model from CarlaSC.}
        \label{fig:kitti-unconditional-generation}
    \end{minipage}\hfill
    \begin{minipage}[t]{0.48\textwidth}
        \centering
        \includegraphics[width=0.95\linewidth]{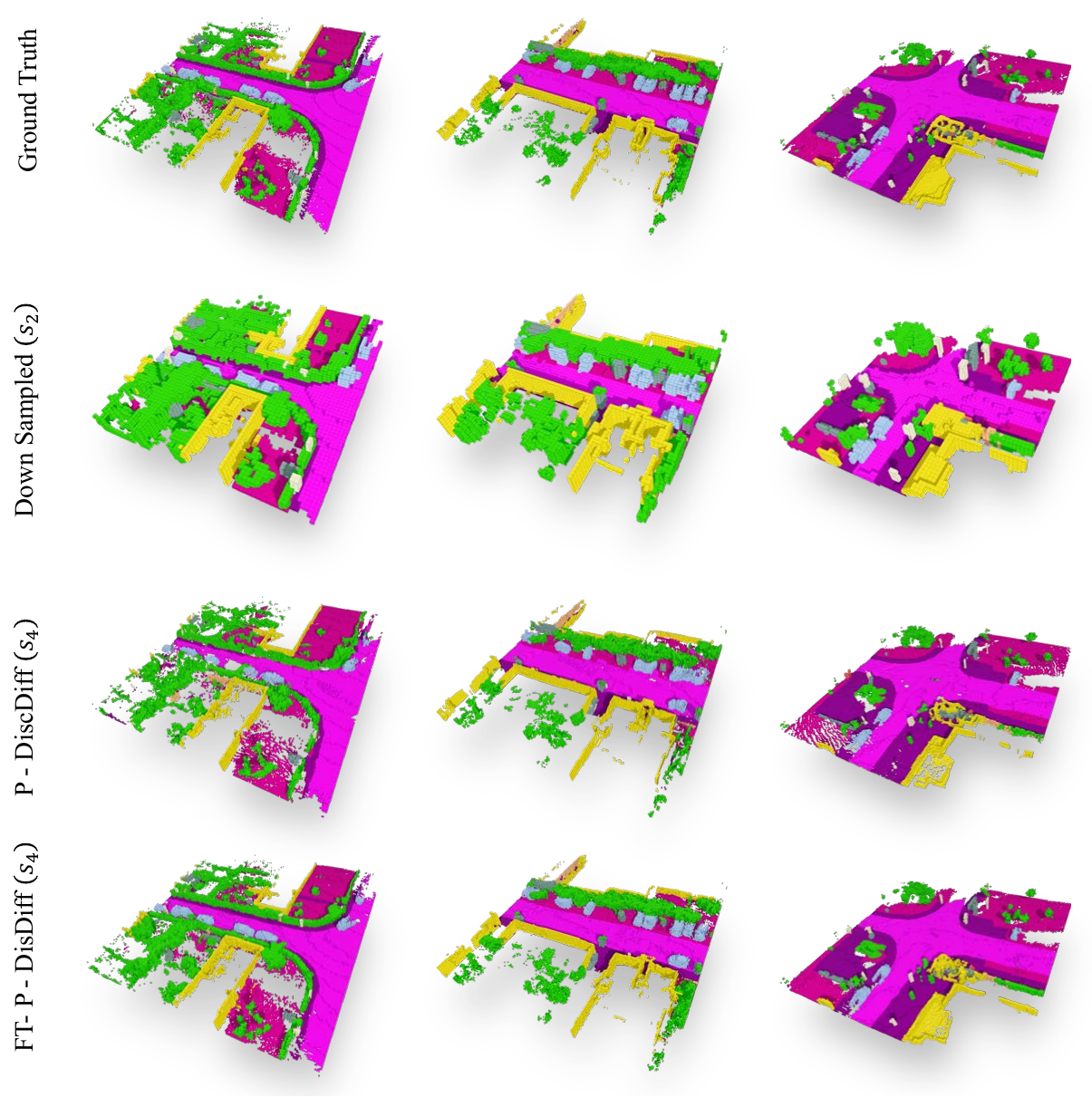}
        \caption{SemanticKITTI conditional generation. \textit{FT} stands for finetuning from CarlaSC models.}
        \label{fig:kitti-conditional-generation}
    \end{minipage}
\end{figure}

\textbf{Cross-dataset.} Figure \ref{fig:kitti-unconditional-generation} and Figure \ref{fig:kitti-conditional-generation} showcase our model's performance on the transferred dataset from CarlaSC to SemanticKITTI for both unconditional and conditional scene generation. The PDD shows enhanced scene quality after finetuning with SemanticKITTI data, as indicated by improved results in Table \ref{semantic-kitti}. This fine-tuning effectively adapts the model to the dataset's complex object distributions and scene dynamics. Notably, despite the higher training effort of the Discrete Diffusion (DD) approach, our method outperforms DD even without fine-tuning, using only coarse scenes from SemanticKITTI. This demonstrates the strong cross-data transfer capability of our approach.

\begin{figure*}[t]
  \centering
  \includegraphics[width = 0.99\textwidth]
{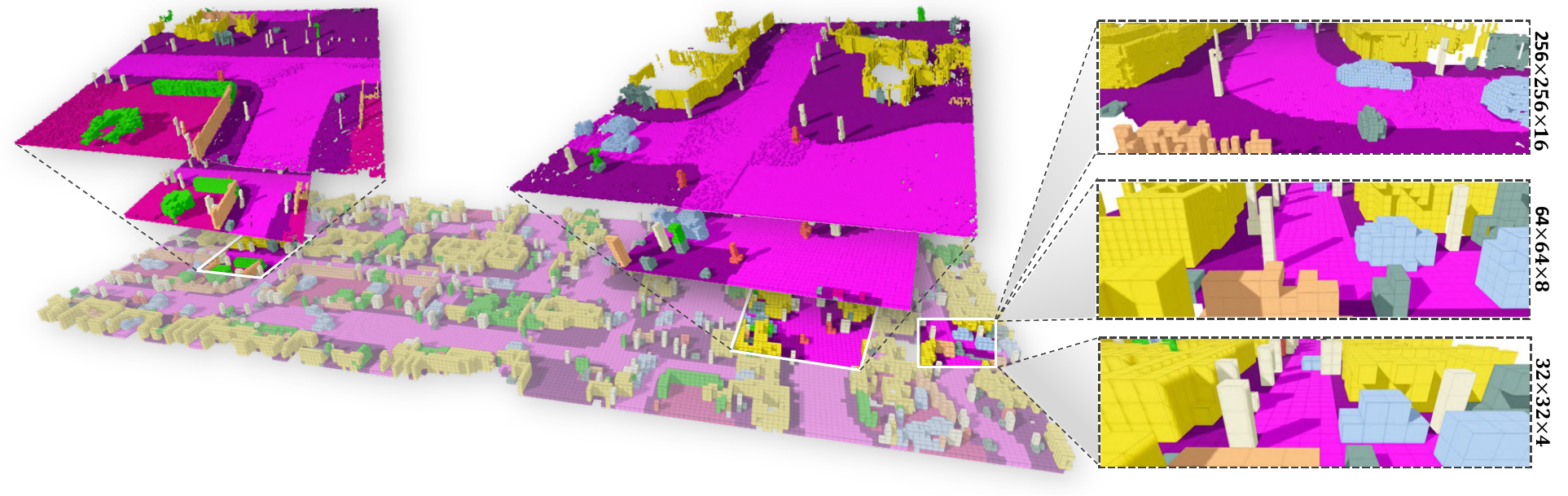}
    \caption{Infinite Scene Generation. Thanks to the pyramid representation, PDD can be readily applied for unbounded scene generation. This involves the initial efficient synthesis of a large-scale coarse 3D scene, followed by subsequent refinement at higher levels.}
    \label{fig:infinity-generation}
\end{figure*}

\begin{table}[t]
\caption{Generation results on SemanticKITTI. \textit{Finetuned Scales} set to None indicates training from scratch and others stand for finetuning corresponding pre-trained CarlaSC model.}
\label{semantic-kitti}
\centering
\resizebox{0.99\textwidth}{!}{
\setlength{\tabcolsep}{5pt}
\begin{tabular}{ccccccc}
\toprule
\multicolumn{1}{c}{\textbf{Method}} & \multicolumn{1}{c}{\textbf{Finetuned Scales}} & \multicolumn{1}{c}{\textbf{Conditioned}} & \multicolumn{1}{c}{\textbf{mIoU(V)}} & \multicolumn{1}{c}{\textbf{mIoU(P)}} & \multicolumn{1}{c}{\textbf{F3D($\downarrow$)}} & \multicolumn{1}{c}{\textbf{MMD($\downarrow$)}} \\
\midrule \midrule
DD \cite{austin2021structured} & $s_4$ & $\times$ & 29.1 & 16.0 & 0.46 & 0.31 \\
\midrule
PDD & None & $\checkmark$ & 33.4 & \textbf{22.8} & \textbf{0.27} & 0.32 \\
PDD & $s_2$, $s_4$ & $\checkmark$ & \textbf{43.9} & \textbf{22.8} & 0.28 & \textbf{0.16} \\
\midrule
PDD & $s_1$ & $\times$ & 31.3 & 23.2 & 0.22 & 0.13 \\
PDD & $s_1$, $s_2$, $s_4$ & $\times$ & \textbf{44.7} & \textbf{24.2} & \textbf{0.21} & \textbf{0.11} \\
\bottomrule
\end{tabular}}
\end{table}

\noindent\textbf{Infinite Scene Generation.}
Figure \ref{fig:infinity-generation} visualizes the process of generating large-scale infinite scenes using our PDD model. As discussed in Section~\ref{sec:3.4}, we first use the small-scale model to swiftly generate a coarse infinite 3D scene (bottom level in Figure \ref{fig:infinity-generation}). We then leverage larger-scale models to progressively add intricate details (middle and top level in Figure \ref{fig:infinity-generation}), enhancing realism. This approach enables our model to produce high-quality, continuous cityscapes without additional inputs, overcoming the limitations of conventional datasets with finite scenes and supporting downstream tasks like 3D scene segmentation.

%% file: sections/06.Conclusion.tex
\section{Conclusion}
In this work, we introduce the Pyramid Discrete Diffusion model (PDD) to address the significant challenges associated with 3D large scene generation, particularly in the context of limited scale and available datasets. The PDD demonstrates a progressive approach to generating high-quality 3D outdoor scenes, transitioning seamlessly from coarse to fine details. Compared to the other methods, the PDD can generate high-quality scenes within limited resource constraints and does not require additional data sources. 
Our extensive experimental results show that PDD consistently performs favourably in both unconditional and conditional generation tasks, establishing itself as a reliable and robust solution for the creation of realistic and intricate scenes.
Furthermore, the proposed PDD method has great potential in efficiently adapting models trained on synthetic data to real-world datasets and suggests a promising solution to the current challenge of limited real-world data. 

%% file: sections/07.Acknowledgements.tex
\section*{Acknowledgement}
Supported by the Intelligence Advanced Research Projects Activity (IARPA) via Department of Interior/ Interior Business Center (DOI/IBC) contract number 140D0423C0074. The U.S. Government is authorized to reproduce and distribute reprints for Governmental purposes notwithstanding any copyright annotation thereon. Disclaimer: The views and conclusions contained herein are those of the authors and should not be interpreted as necessarily representing the official policies or endorsements, either expressed or implied, of IARPA, DOI/IBC, or the U.S. Government. Y. Liu and C. Li are supported in part by the National Science Foundation of China (Grant No: 62202395).

%% file: sections/X_suppl.tex
\clearpage
\renewcommand\thesection{\Alph{section}}
\renewcommand\thetable{\alph{table}}
\renewcommand\thefigure{\alph{figure}}
\title{Pyramid Diffusion for Fine 3D Large Scene Generation - Supplementary Material} 
\author{}
\titlerunning{Pyramid Diffusion for Fine 3D Large Scene Generation}
\authorrunning{Y.~Liu et al.}
\institute{}
\maketitle

This supplementary document details our evaluation setup, hyperparameters setting, data pre-processing, and more experimental results. The supplementary material is organized as follows:

\begin{compactitem}
\setlength{\itemsep}{1pt}
\setlength{\parsep}{1pt}
\setlength{\parskip}{1pt}
\item {Evaluation Setup in Section \ref{sec:evaluation-setup}.}
\item{Hyperparameters Setting in Section \ref{sec:hyperparameters-setting}.}
\item{Data Pre-processing in Section \ref{sec:datasets:pre-processing}.}
\item{Additional Experimental Results in Section \ref{sec:additional-exp-results}.}
\item{Additional Discussion in Section \ref{sec:addition-discussion}.}

\end{compactitem}

\section{Evaluation Setup}
\label{sec:evaluation-setup}
In all evaluations of our paper, we consistently randomly sample 1,000 distinct scenes to assess the generation quality. The specific methodologies for F3D (Fréchet 3D Distance) and MMD (Maximum Mean Discrepancy) are as follows.
\subsection{F3D}
It is an evaluation metric adapted from the 2D Fréchet Inception Distance (FID)~\cite{heusel2017gans} to evaluate the quality of generated 3D scenes. Implementing F3D aims to complement semantic segmentation by capturing the richness and diversity of generated scenes, which semantic segmentation might overlook. F3D ensures that the generated scenes maintain complexity and reflect the similarity between generated 3D scenes and real-world structures.

Our F3D is calculated following the next steps. Initially, we pre-train a 3D CNN-based autoencoder, which is subsequently utilized to extract high-dimensional features from the generated 3D scenes. The F3D is then computed akin to FID, leveraging the extracted features to evaluate the discrepancies between the generated and real scenes. Mathematically, F3D is represented as: 
\begin{equation}
\ \text{F3D} = \| \boldsymbol{\mu}_g - \boldsymbol{\mu}_r\|^2 + \mathbf{Tr} \left( \mathbf{\Sigma}_g + \mathbf{\Sigma}_r - 2 (\mathbf{\Sigma}_g \mathbf{\Sigma}_r)^{1/2} \right)
\end{equation}
where $\boldsymbol{\mu}_g$, $\boldsymbol{\mu}_r$ are the feature means, and $\mathbf{\Sigma}_g$, $\mathbf{\Sigma}_r$ are the feature covariances of the generated and real scenes.

\subsection{MMD}
We incorporate the Maximum Mean Discrepancy (MMD) as a key statistical measure to quantify the disparity between the distributions of generated and real-world scenes. Following a method akin to our F3D approach, we initially extract high-dimensional features from the 3D scenes using a 3D CNN architecture which is used in F3D. Subsequently, we employ a Gaussian kernel, expressed as 
\begin{equation}
k(\boldsymbol{f}, \boldsymbol{f}') = \exp\left(-\frac{\|\boldsymbol{f} - \boldsymbol{f}'\|^2}{2\sigma^2}\right) 
\end{equation}
where $\sigma$ is the kernel width to map these features into a higher-dimensional space for MMD calculation. The bandwidth $\sigma$ is determined using the median heuristic, a robust method to estimate the scale of data in the feature space. The MMD formula is given by 

\begin{equation}
\textbf{MMD}^2 = \left\| \frac{1}{n} \sum_{i=1}^n k(\boldsymbol{f}_i, \cdot) - \frac{1}{m} \sum_{j=1}^m k(\boldsymbol{f}'_j, \cdot) \right\|^2 
\end{equation}
$\boldsymbol{f}$ and $\boldsymbol{f}'$ are the extracted features from the generated and real dataset, respectively. The utilization of MMD, especially with the Gaussian kernel, not only captures the overall statistical distribution but also considers finer details in the feature space, rendering it an indispensable tool in our evaluation protocol.

\section{Hyperparameters Setting}
\label{sec:hyperparameters-setting}
\subsection{Pyramid Discrete Diffusion Models}
\noindent\textbf{Training Hyperparameters. }In the main experiment of our Pyramid Discrete Diffusion, a total of four diffusion models, namely PDD ($s_1$), PDD ($s_2$), PDD ($s_3$), and PDD ($s_4$), are utilized. Each model is trained on four NVIDIA A100 GPUs with batch sizes set to 128, 32, 16, and 8, respectively. A unified learning rate of $10^{-3}$ is applied and the AdamW optimizer is used to train each model for 800 epochs. Additionally, during training, data augmentation techniques, including flipping and rotation, are employed to enhance the robustness of the models.

\noindent\textbf{Cross-dataset Hyperparameters. }In our paper, we demonstrate the capability of our method for cross-dataset generation. The models labeled with \textit{FT} are those where the generation model trains on the CarlaSC dataset~\cite{wilson2022motionsc} and undergoes finetuning using the SemanticKITTI dataset~\cite{behley2019semantickitti}. Specifically, we use the PDD ($s_4$) model for finetuning, which has already been trained for 800 epochs on CarlaSC. This model is then further trained for 200 epochs using the SemanticKITTI dataset, with all other hyperparameters remaining unchanged. Consequently, the entire model completes a total of 1,000 epochs of training.


\noindent\textbf{Sampling Hyperparameters. }In our paper, all the sampled scenes we demonstrate are generated using 100 diffusion steps.

\subsection{Evaluation Models}
We train six different models across three distinct networks in our evaluation process. These include 3D CNNs for evaluating Fréchet 3D Distance (F3D) and Maximum Mean Discrepancy (MMD), SparseUNet~\cite{graham20183d}, and PointNet$++$~\cite{qi2017pointnet++} for semantic segmentation evaluation.


\noindent\textbf{3D CNN Hyperparameters. } As mentioned in Section~\ref{sec:evaluation-setup} of our paper, the evaluation of F3D and MMD relies on a well-trained 3D CNN network. We train two separate 3D CNNs for the CarlaSC and SemanticKITTI datasets to perform their respective F3D and MMD evaluations but maintain identical training parameters for both networks. The 3D CNNs are trained as an autoencoder, with the primary aim of feature extraction. The loss function employed in this training is a balanced cross-entropy, a form of reconstruction loss. We set the batch size to 16 and employed SGD as the optimizer with a cosine scheduler. The networks are trained for 30 epochs at a $10^{-2}$ learning rate. We adhere to the original dataset splits for training, using the training sets as delineated in the original dataset distributions.


\noindent\textbf{SparseUNet Hyperparameters. }SparseUNet~\cite{graham20183d}, designed for voxel-based semantic segmentation, is trained with two separate models for the CarlaSC and SemanticKITTI datasets. All training parameters of these SparseUNet models are consistent with those of the 3D CNNs. This includes a batch size 16, utilizing SGD as the optimizer, employing a cosine scheduler, and training for 30 epochs at a learning rate of $10^{-2}$. The training adheres to the original dataset divisions for the respective training sets.


\noindent\textbf{PointNet$++$ Hyperparameters. }Similarly to our previous approach, we train a separate semantic segmentation model based on PointNet$++$ (SSG) for each dataset. Each model is trained with a batch size of 16. We randomly sample 16,384 points for each scene, and if the number of points is insufficient, we supplement them by replication. We set the optimizer to Adam and use a step scheduler with a step size 5. The models are trained for 30 epochs at a learning rate of $5\times 10^{-3}$.

\subsection{Baseline}
\noindent\textbf{Unconditional Generation. }In our paper's unconditional generation comparison experiments, we include two baseline models: the Discrete Diffusion model (DD)~\cite{austin2021structured} and the Latent Diffusion model~\cite{lee2023diffusion}. For the DD, following the descriptions in~\cite{lee2023diffusion}, we train generation models with a scale of $256\times256\times16$ for both the CarlaSC and SemanticKITTI datasets. Regarding the Latent Diffusion model, we utilize the pre-trained model published in the work~\cite{lee2023diffusion} as our baseline. However, since the original work trains the model at a scale of $128\times128\times8$, during the generation phase, we use the Scale Adaptive Function to upsample the generated scenes to match our scale of $256\times256\times16$.


\noindent\textbf{Conditional Generation. }In our paper's conditional generation experiment sections, we use a discrete diffusion model conditioned on point clouds as our baseline. The scene is divided into a voxel-based representation based on scale, and each voxel is assigned a binary value (0 or 1). A voxel is set to 1, indicating the presence of points if it contains one or more points from the point cloud.

We recognize that this comparison might not be entirely fair due to scale differences. However, the final upscaling and subsequent visualization of the generated scenes allow for a clear discernment of quality differences. It's important to note that training in the original method at the larger scale of $256\times256\times16$ would require substantial computational resources, specifically over 16 days on 4 A100 GPUs. Despite this potential bias, the visual evaluation method, after upscaling, effectively demonstrates the quality distinction between the generated scenes, which is crucial for our evaluation. This approach is chosen as the most feasible solution given our resource constraints.

\section{Datasets Pre-processing}
\label{sec:datasets:pre-processing}
Our paper's experiments utilize two outdoor scene datasets: CarlaSC~\cite{wilson2022motionsc} and SemanticKITTI~\cite{behley2019semantickitti}. CarlaSC is a dataset collected through simulated road scenes and is primarily used in our main experiments. On the other hand, SemanticKITTI, gathered from real-world scenes, is employed in experiments focusing on cross-dataset applications. Due to the different origins of these datasets and their varied labels, we undertake specific processing steps to facilitate better experimentation. The details of these processing steps are as follows.

\subsection{CarlaSC}

\begin{table}[tp]
\centering
\caption{\textbf{Simplified Semantic Labels for the CarlaSC Dataset After Merging.} The table lists the 10 consolidated classes used in our experiments, with \textit{0} denoting unclassified elements not shown here.}
\label{merged-labels}
\begin{tabular}{cc|ccl}
\toprule
\textbf{Index} & \textbf{Label}      & \textbf{Index} & \textbf{Label}      &  \\ \cline{1-4}
\hline
\hline
1     & Building   & 6     & Road       &  \\
2     & Fences     & 7     & Ground     &  \\
3     & Other      & 8     & Sidewalk   &  \\
4     & Pedestrian & 9     & Vegetation &  \\
5     & Pole       & 10    & Vehicle    &  \\
\bottomrule
\end{tabular}
\end{table}

CarlaSC~\cite{wilson2022motionsc}, primarily employed in our main experiments, is a synthetic dataset featuring outdoor road point cloud scenes. Originally comprising 23 semantic labels, these merge according to the dataset's official guidelines to simplify the categorization process. The dataset encompasses 11 semantic classes, as detailed in Table \ref{merged-labels}, with \textit{0} representing the \textit{unclassified} category. This dataset includes 18 scenes for training, 3 for validation, and 3 for testing. In our experiments, we utilize a high-scale version of CarlaSC, where each scene has a scale of $256\times256\times16$ voxels, covering a physical space of 25.6 meters both in front of and behind the radar scanner and extending up to a height of 3 meters.

\begin{table*}[t]
\begin{center}
\caption{Conversion of SemanticKITTI Labels to Correspond with CarlaSC's 11 Categories. Labels listed as \textit{remove} are absent in CarlaSC, while those marked with \textit{-} are omitted from semantic segmentation according to the original settings.}
\label{semantic-kitti-label-projection}
\scriptsize{
\resizebox{0.99\linewidth}{!}{
\begin{tabular}{ccccc|ccccc}
\toprule
Index & Original Labels & Mapped index & Mapped Labels & Ratio    & Index & Original Labels     & Mapped index & Mapped Labels & Ratio   \\
\hline
\hline
0                & unlabeled& 0            & Unclassified   & $1.8 \times 10^{-2}$& 51             & fence               & 2            & fence         & $7 \times 10^{-2}$\\
1                & outlier& -& -             & $2 \times 10^{-4}$& 52             & other-structure     & 3            & other         & $2 \times 10^{-3}$\\
10               & car             & 10           & vehicle       & $4.1 \times 10^{-2}$& 60             & lane-marking        & 6            & road          & $4 \times 10^{-5}$\\
11               & bicycle         & remove       & -             & $1 \times 10^{-4}$& 70             & vegetation          & 9            & vegetation    & $2.6 \times 10^{-1}$\\
13               & bus             & remove       & -             & $3 \times 10^{-5}$& 71             & trunk               & 9            & vegetation    & $6 \times 10^{-3}$\\
15               & motorcycle      & remove       & -             & $3 \times 10^{-4}$& 72             & terrain             & 7            & ground        & $8 \times 10^{-2}$\\
16               & on-rails        & remove       & -             & 0        & 80             & pole                & 5            & pole          & $3 \times 10^
{-3}$\\
18               & truck           & remove       & -             & $2 \times 10^{-3}$& 81             & traffic-sign        & 5            & pole          & $6 \times 10^{-4}$\\
20               & other-vehicle   & remove       & -             & $2 \times 10^{-3}$& 99             & other-object        & 3            & other         & $1 \times 10^{-2}$\\
30               & person          & 4            & pedestrian    & $2 \times 10^{-4}$& 252            & moving-car          & -& -             & $2 \times 10^
{-3}$\\
31               & bicyclist       & remove       & -             & $1 \times 10^{-8}$& 253            & moving-bicyclist    & -& -             & $1 \times 10^{-4}$\\
32               & motorcyclist    & remove       & -             & $5 \times 10^{-9}$& 254            & moving-person       & -& -             & $2 \times 10^{-4}$\\
40               & road            & 6            & road          & $2 \times 10^{-1}$& 255            & moving-motorcyclist & -& -             & $3 \times 10^{-5}$\\
44               & parking         & 7            & ground        & $1.5 \times 10^{-2}$& 256            & moving-on-rails     & -& -             & 0\\
48               & sidewalk        & 8            & sidewalk      & $1.4 \times 10^{-1}$& 257            & moving-bus          & -& -             & $1 \times 10^{-4}$\\
49               & other-ground    & 7            & ground        & $4 \times 10^{-3}$& 258            & moving-truck        & -& -             & $1 \times 10^{-4}$\\
50               & building        & 1            & building      & $1.3 \times 10^{-1}$& 259            & moving-other        & -& -             & $4 \times 10^{-5}$\\
\bottomrule
\end{tabular}
}
}
\end{center}
\end{table*}

\subsection{SemanticKITTI}

SemanticKITTI~\cite{behley2019semantickitti}, utilized for cross-dataset validation, showcases diverse environments, including inner-city traffic, residential areas, highways, and countryside roads. After voxelization, it comprises 22 sequences: sequences 00 to 10 (excluding 08) for training, 08 for validation, and 11 to 20 for testing. However, due to the absence of semantic labels for the test set in the official SemanticKITTI release, our assessment of scene generation depends on the validation set. The chosen scope extends 51.2 meters ahead of the vehicle, 25.6 meters laterally, and 6.4 meters in height, yielding a voxel scale of \(256 \times 256 \times 32\). Originally featuring 28 categories, SemanticKITTI undergoes a class remapping or removal process to align with the 11 categories found in CarlaSC, detailed in Table~\ref{semantic-kitti-label-projection}. 

In this alignment, we have removed certain categories from SemanticKITTI. This decision is primarily based on the original dataset's configuration for semantic segmentation, such as moving objects, and the absence of corresponding labels in the CarlaSC dataset, like bicycles and motorcycles. Notably, the removed categories represent only a minor portion of the total number of labels.
This ensures the labels conform to the relationships in Table~\ref{merged-labels}. Additionally, we omit the upper 16 voxels in height from SemanticKITTI to synchronize with CarlaSC's height representation, a decision justified by the rarity of objects above this limit and the need for consistency in dataset comparison.

\section{Additional Experiment Results}
\label{sec:additional-exp-results}
Due to the limitations on the length of the main text, we have reserved the primary experimental results for inclusion in the main body of the paper. In this section, we provide supplementary experimental outcomes not mentioned in the main text and more showcases of our visualization effects.

\subsection{Generation Quality}
\begin{figure*}
  \centering
  \includegraphics[width = 0.99\linewidth]
  {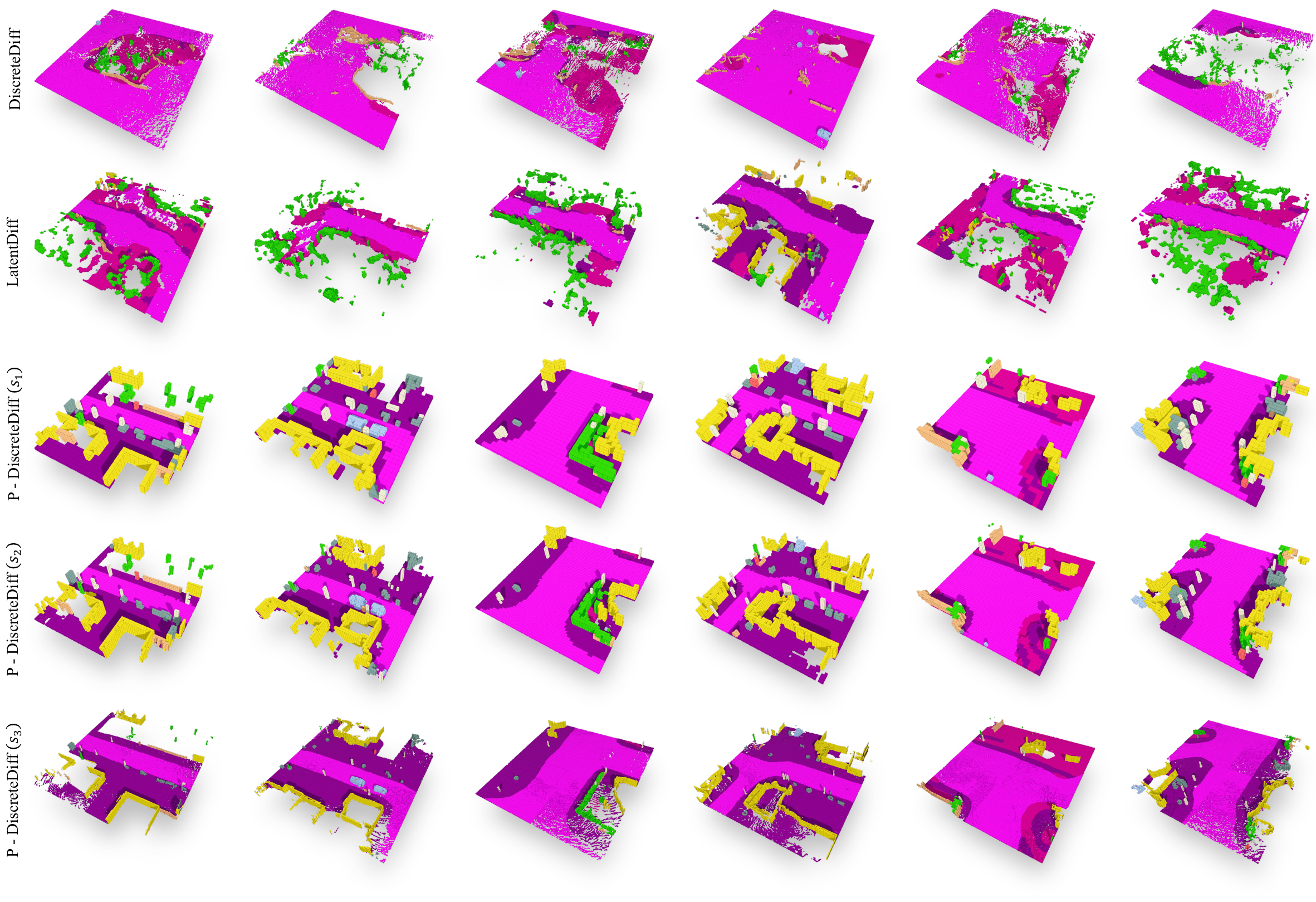}
    \caption{\textbf{Additional visualization of unconditional generation results on CarlaSC.} Our method produces more diverse scenes compared to the two baseline models \cite{lee2023diffusion, austin2021structured}.}
    \label{fig:supp-uncon-carla-1}
\end{figure*}

\begin{figure*} 
  \centering
  \includegraphics[width = 0.99\linewidth]
  {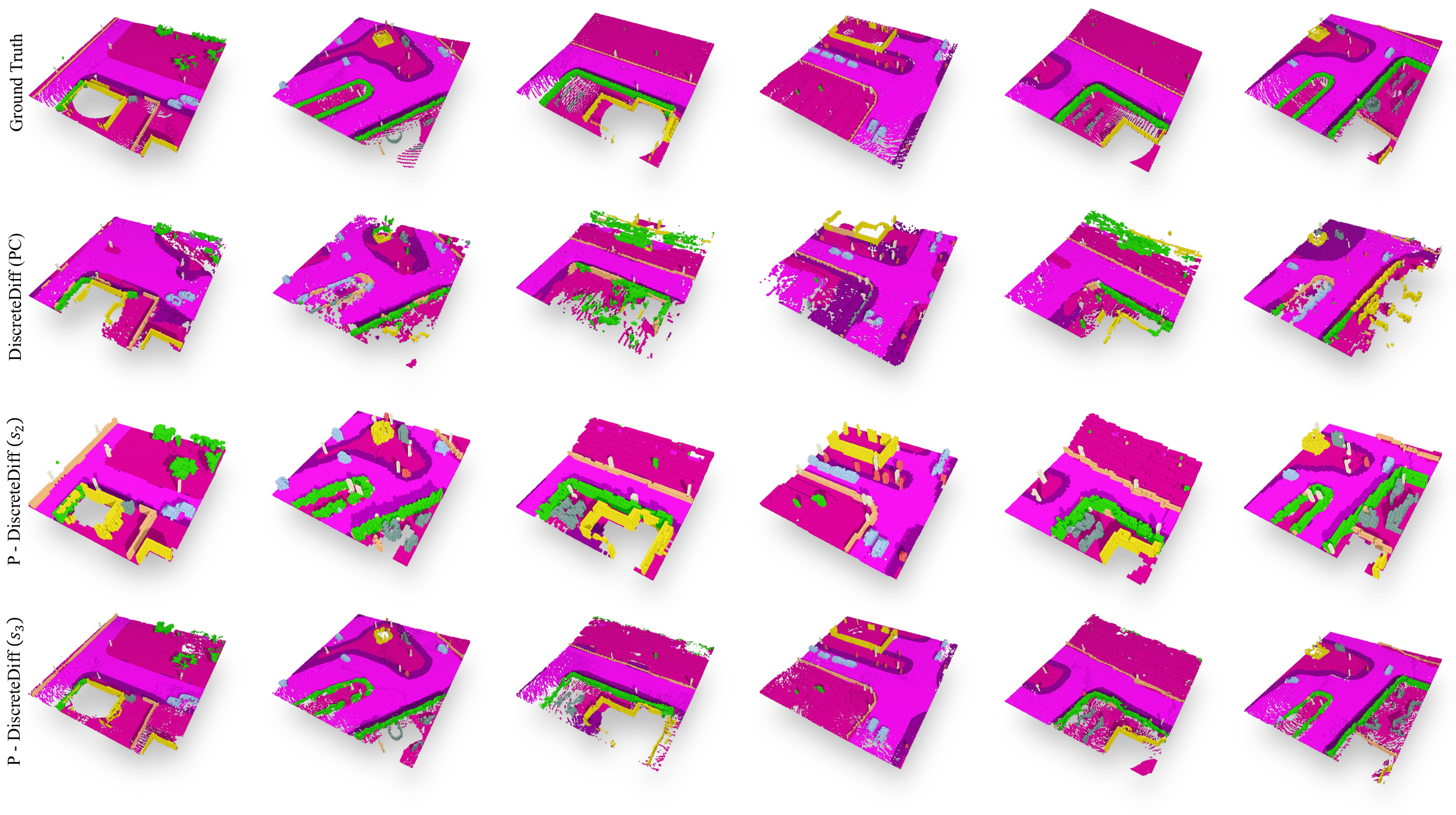}
    \caption{\textbf{Additional visualization of conditional generation results on CarlaSC.} \textit{PC} stands for point cloud condition.}
    \label{fig:supp-uncon-carla-4}
\end{figure*}

\begin{figure*}
  \centering
  \includegraphics[width = 0.99\linewidth]
  {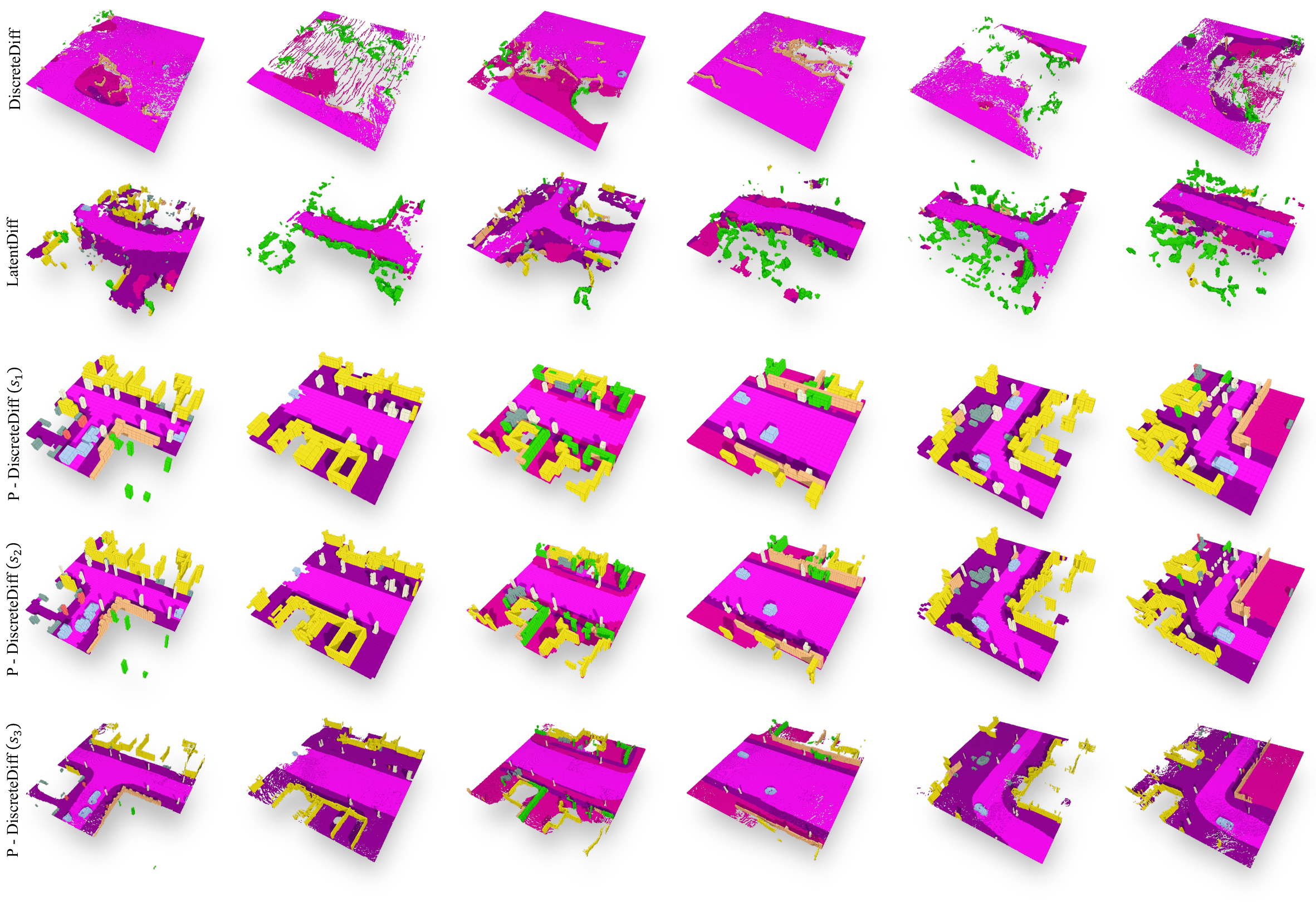}
    \caption{\textbf{Additional visualization of unconditional generation results on CarlaSC.} Our method produces more diverse scenes compared to the two baseline models \cite{lee2023diffusion, austin2021structured}.}
    \label{fig:supp-uncon-carla-2}
\end{figure*}

\begin{figure*}
  \centering
  \includegraphics[width = 0.99\linewidth]
  {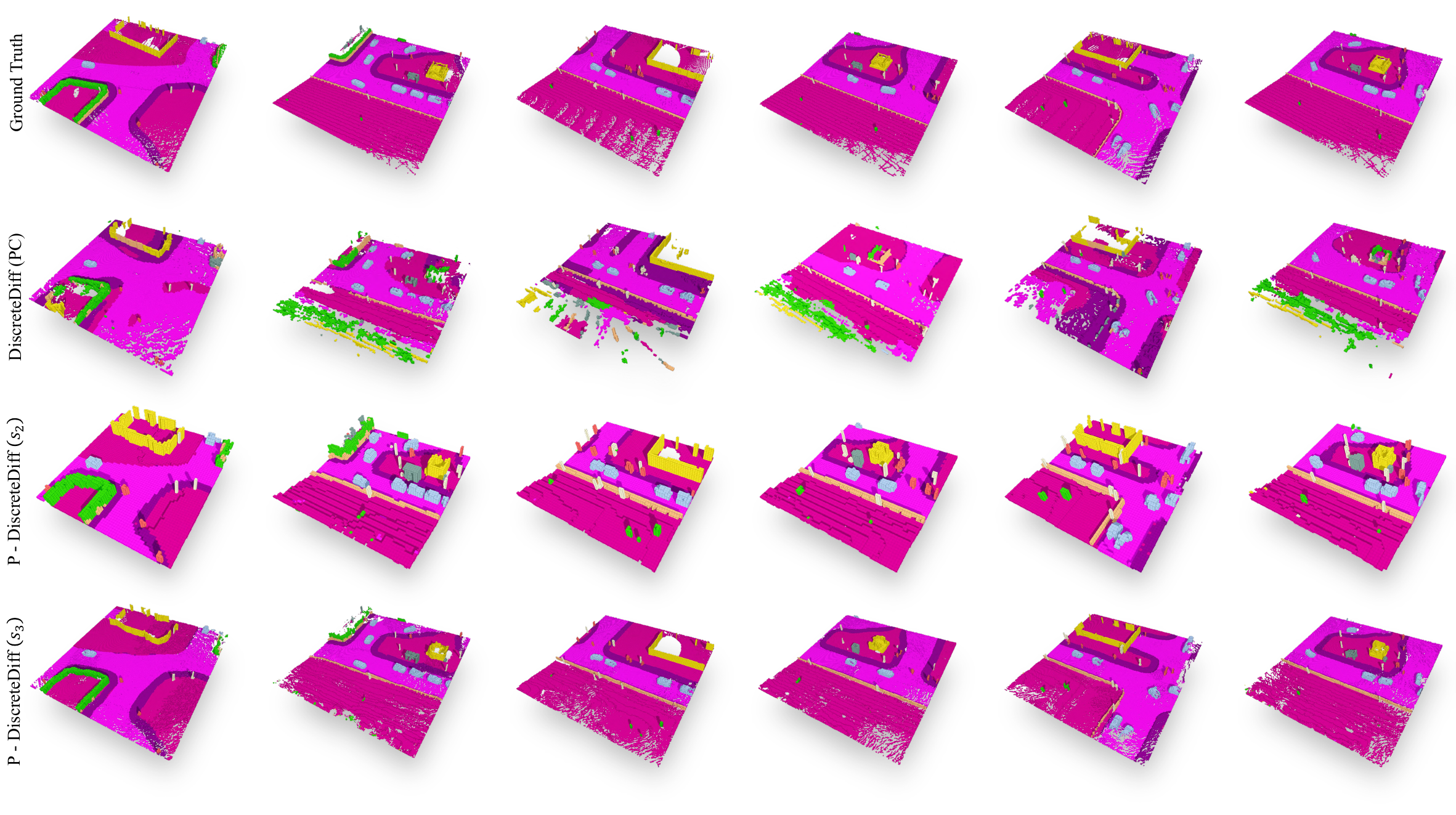}
    \caption{\textbf{Additional visualization of conditional generation results on CarlaSC.} \textit{PC} stands for point cloud condition.}
    \label{fig:supp-uncon-carla-3}
\end{figure*}

In this section, we present an array of qualitative comparative results. Figures \ref{fig:supp-uncon-carla-1} and \ref{fig:supp-uncon-carla-2} showcase the outcomes of our \textit{Unconditional Generation}, from which we can discern that scenes generated using the PDD method achieve greater semantic accuracy and encompass more details within the generated scenes. Figures~\ref{fig:supp-uncon-carla-3} and~\ref{fig:supp-uncon-carla-4} illustrate the results of our \textit{Conditional Generation}. The outcomes reveal that the scenes generated by our conditional generation method closely resemble the ground truth. In contrast, while the approach using point clouds as a condition for scene restoration maintains structural similarity, it exhibits numerous inaccuracies in label correctness.

\subsection{Applications}
\begin{figure}[htbp]
    \begin{minipage}[t]{0.48\linewidth}
        \centering
        \includegraphics[width=0.99\linewidth]{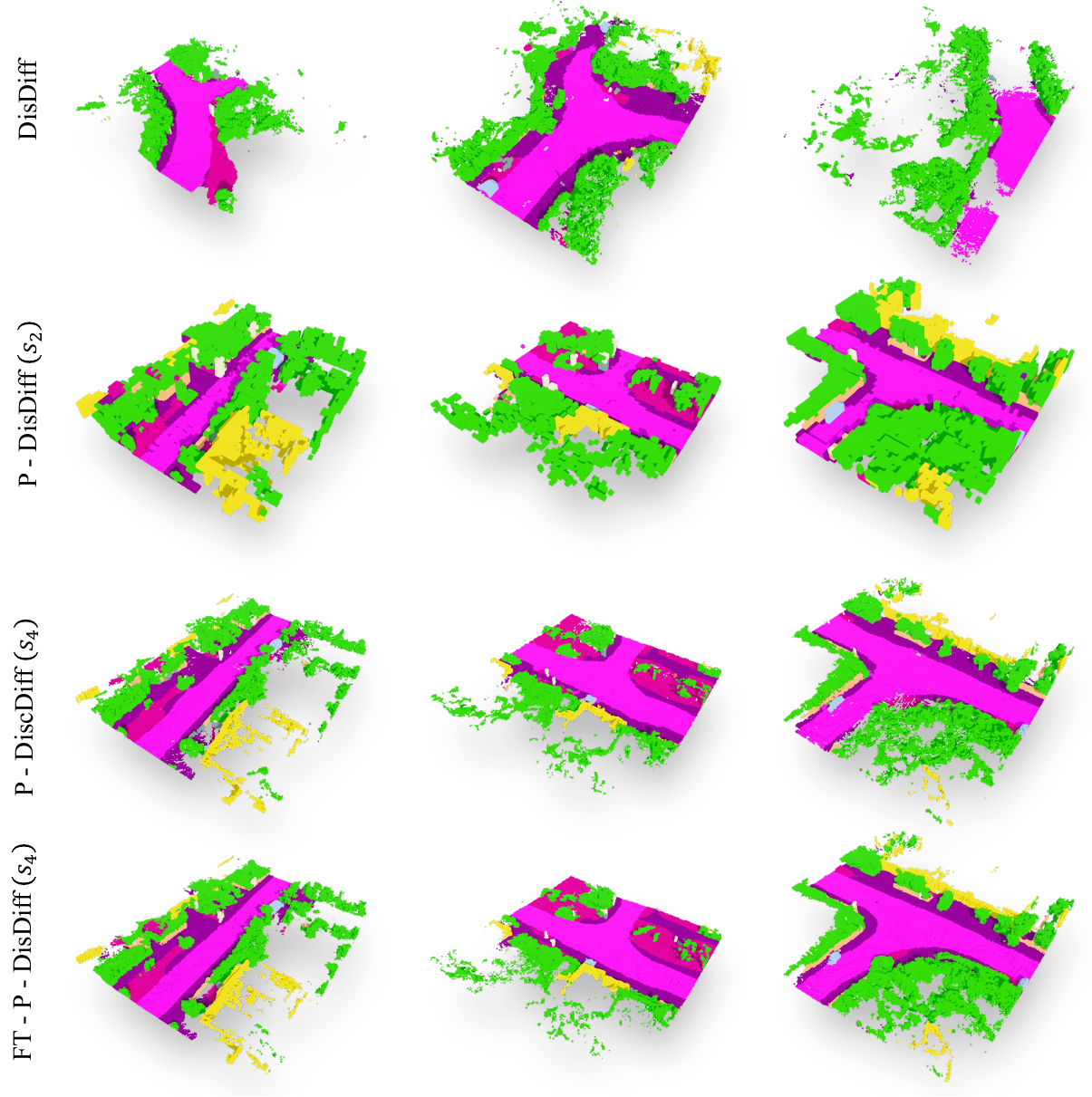}
        \caption{\textbf{Additional SemanticKITTI unconditional generation.} \textit{FT} stands for finetuning pre-trained model from CarlaSC.}
        \label{fig:supp-kitti-uncon-1}
    \end{minipage}\hfill
    \begin{minipage}[t]{0.48\linewidth}
        \centering
        \includegraphics[width=0.99\linewidth]{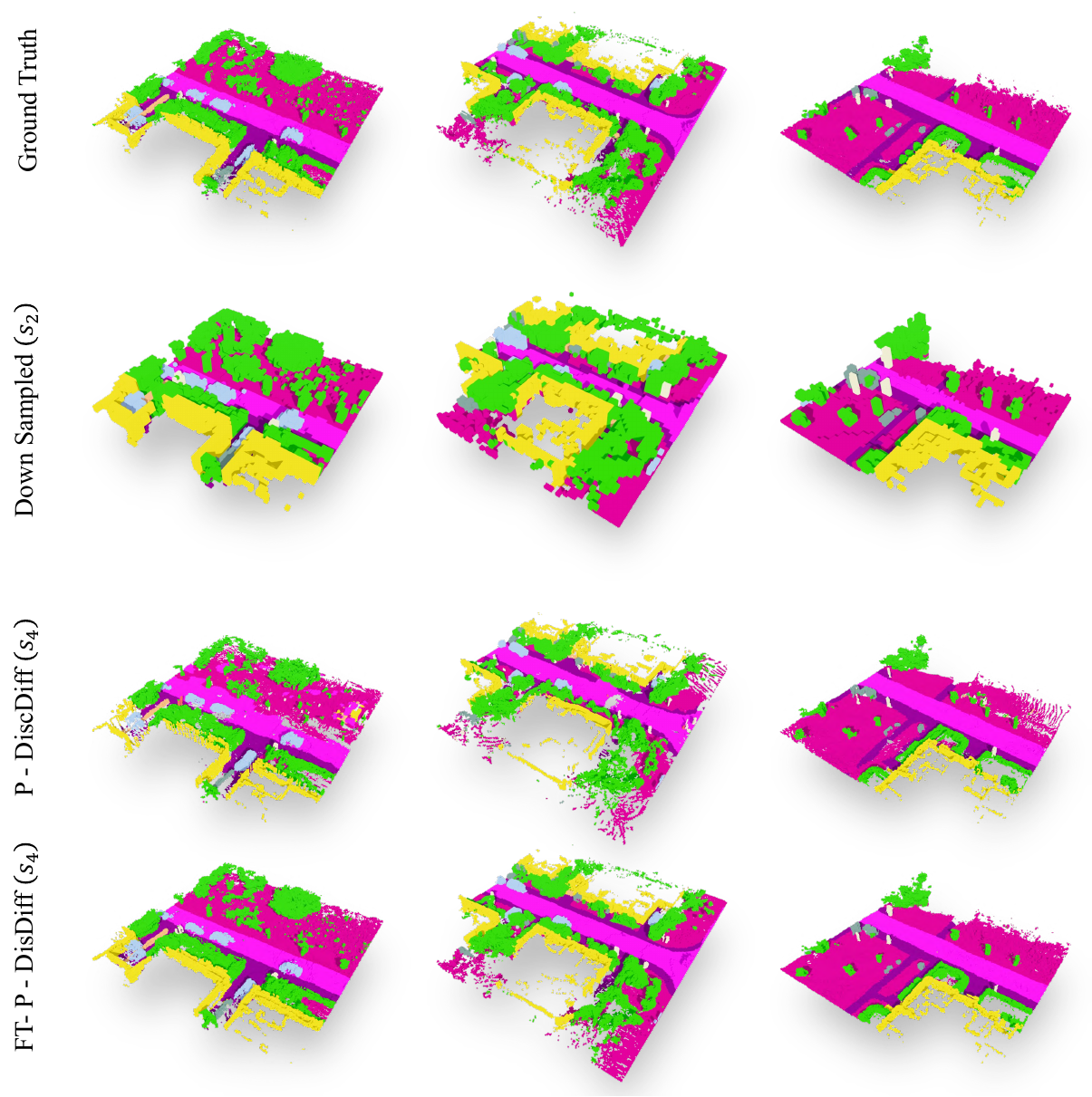}
        \caption{\textbf{Additional SemanticKITTI conditional generation.} Our proposed PDD achieves results close to the groundtruth. Note that \textit{FT} stands for finetuning from CarlaSC models.}
        \label{fig:supp-kitti-con-1}
    \end{minipage}
\end{figure}

\begin{figure}[htbp]
    \begin{minipage}[t]{0.48\linewidth}
        \centering
        \includegraphics[width=0.99\linewidth]{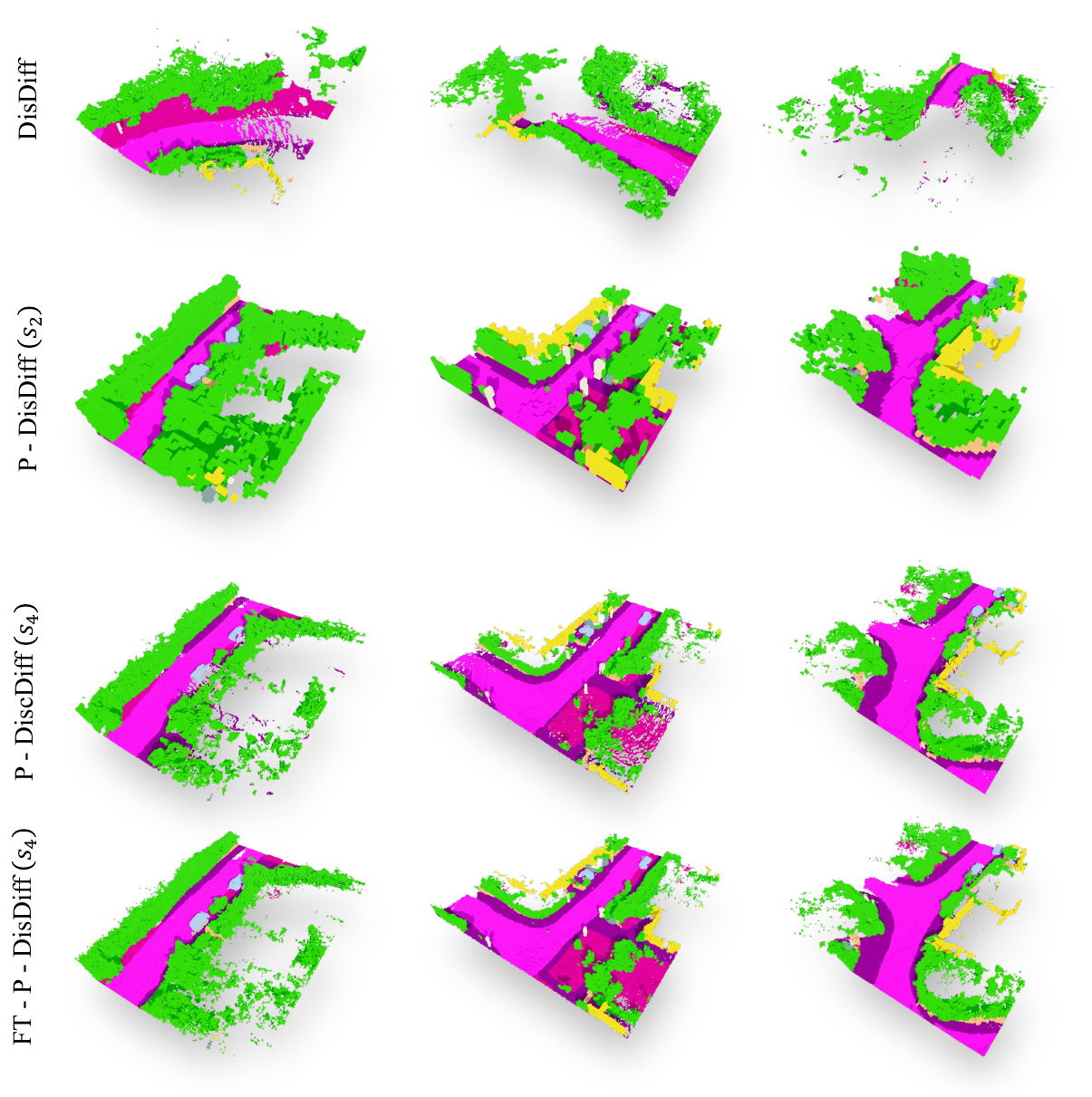}
        \caption{\textbf{Additional SemanticKITTI unconditional generation.} \textit{FT} stands for finetuning pre-trained model from CarlaSC.}
        \label{fig:supp-kitti-uncon-2}
    \end{minipage}\hfill
    \begin{minipage}[t]{0.48\linewidth}
        \centering
        \includegraphics[width=0.99\linewidth]{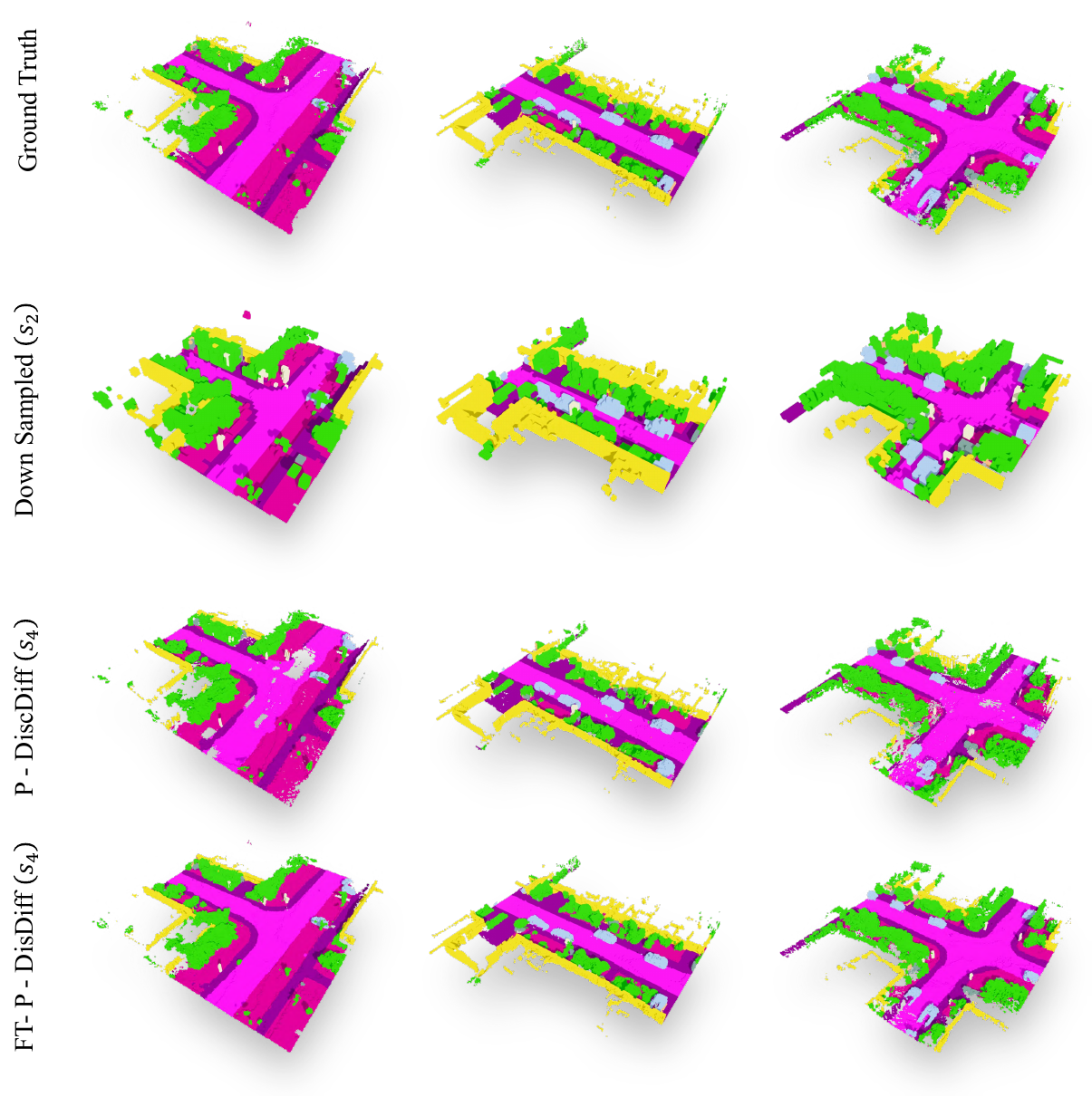}
        \caption{\textbf{Additional SemanticKITTI conditional generation.} Our proposed PDD achieves results close to the groundtruth. Note that \textit{FT} stands for finetuning from CarlaSC models.}
        \label{fig:supp-kitti-con-2}
    \end{minipage}
\end{figure}

Our PDD method extends to two applications: cross-dataset and infinite scene generation. Figures~\ref{fig:supp-kitti-uncon-1},~\ref{fig:supp-kitti-uncon-2},~\ref{fig:supp-kitti-con-1} and~\ref{fig:supp-kitti-con-2} provide additional visualizations of cross-dataset scene generation on the SemanticKITTI dataset. Specifically, Figure~\ref{fig:supp-kitti-uncon-1} and~\ref{fig:supp-kitti-uncon-2} illustrate the distinctions in generation effects between DD and our method. Our method consistently generates more coherent semantic scenes, encompassing a wider range of objects and richer details. For Figure~\ref{fig:supp-kitti-con-1} and~\ref{fig:supp-kitti-con-2}, we initially downsample the ground truth to $64 \times 64 \times 8$ and then restore it to $256 \times 256 \times 16$. These results demonstrate our method's capability to reconstruct scenes closely aligned with the ground truth accurately. 

\begin{figure*}
  \centering
  \includegraphics[width = 0.90\linewidth]
  {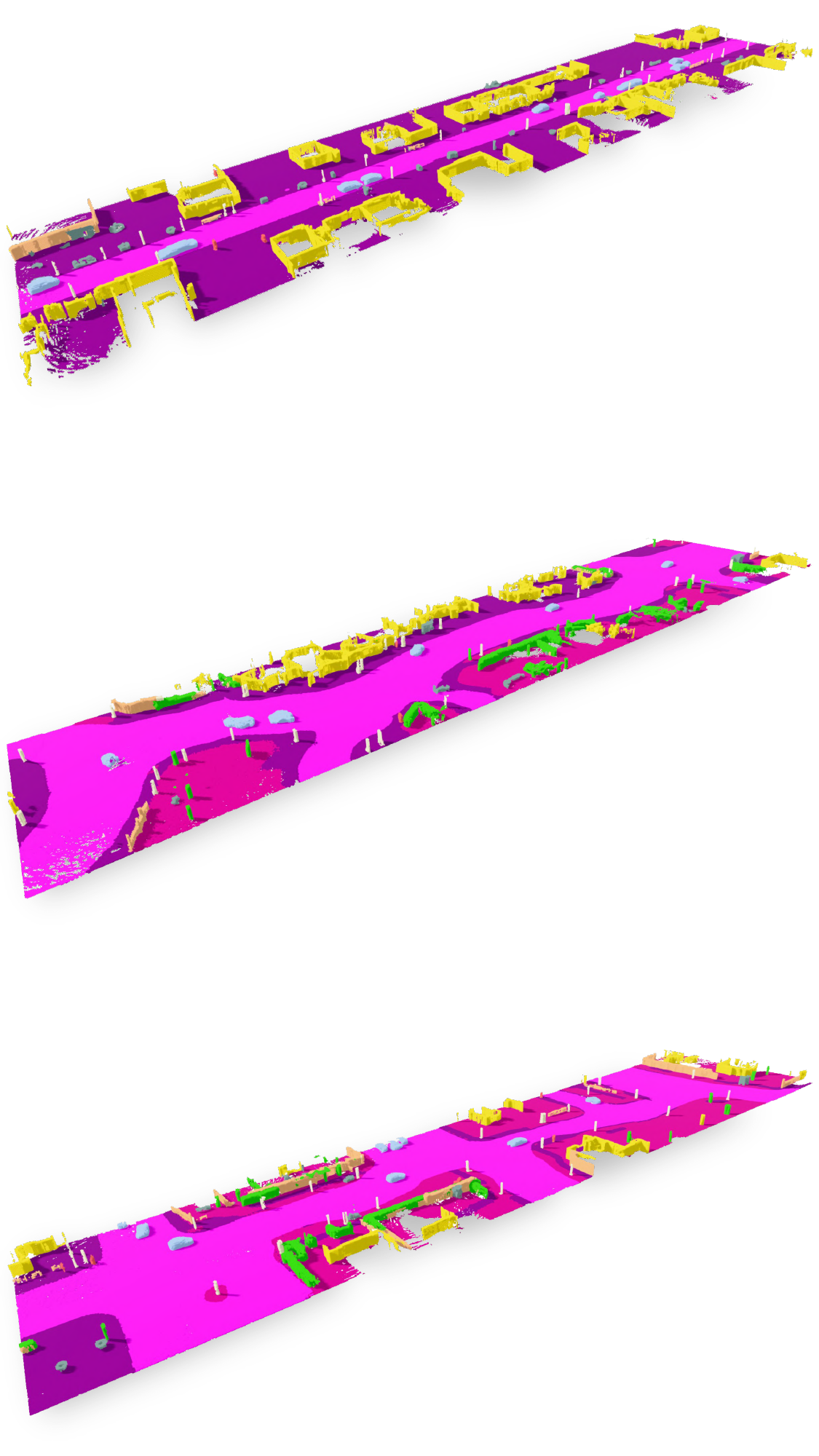}
    \caption{\textbf{Infinite Scene Generation.} Using PDD, we generate three different scenes. Our method produces infinite and consistent urban landscapes, seamlessly blending diverse urban elements to create a coherent and realistic cityscape.}
    \label{fig:supp-infinity}
\end{figure*}

Additionally, Figure \ref{fig:supp-infinity} demonstrates the capability of our method to generate infinite scenes. 
Overall, the scenes generated by our method show high diversity with minimal repetition. We note certain artifacts, such as road interruptions, which can be attributed to two main factors.
First, our approach, in line with prior works~\cite{lin2023infinicity}, assumes a local dependency for infinite scene generation (see Equation 6 in the paper). 
Though efficient, this approach may result in artifacts due to its tendency to overlook long-range dependencies. This could be addressed by sequential models like transformers, which is beyond the scope of this paper. 
Furthermore, the training dataset~\cite{wilson2022motionsc, behley2019semantickitti} consists of radar-scanned road segments rather than complete regional scenes, which potentially causes artifacts.
The extrapolation beyond the scanned scene segments poses a significant challenge, resulting in disruptions within the generated infinite scenes.
We anticipate that, with enhanced datasets and the incorporation of sequential models, our method has the potential to generate better infinite 3D scenes.

\subsection{Ablation Study}
\noindent\textbf{Pyramid Diffusion. }Due to space constraints, only a subset of experimental results are provided in Table 3. We now present the full experimental outcomes in Table \ref{complete-pyramid-diffusion}. The full results show that the performance for both conditional and unconditional generation improves incrementally with adding scales. However, within the four-scale pyramid, the increase is only marginal.

\begin{table}[tp]
\centering
\caption{Complete comparison of different diffusion pyramids on 3D semantic scene generation.}
\label{complete-pyramid-diffusion}
\resizebox{0.99\linewidth}{!}{
\setlength{\tabcolsep}{5pt}
\begin{tabular}{cccccccc}
\toprule
\multirow{1}{*}{\textbf{Cascaded}}& \multicolumn{1}{c}{\multirow{1}{*}{\textbf{Condition}}} & \multirow{1}{*}{\textbf{mIoU(V)}}& \multirow{1}{*}{\textbf{MA(V)}}& \multirow{1}{*}{\textbf{mIoU(P)}}& \multirow{1}{*}{\textbf{MA(P)}}& \multirow{1}{*}{\textbf{F3D$(\downarrow)$}}& \multirow{1}{*}{\textbf{MMD$(\downarrow)$}}\\ \hline \hline
$s_4$& $\times$                                       & 40.0& 63.7& 25.5& 38.7& 1.36                 & 0.60\\
$s_1 \rightarrow s_4$& $\times$                                       & 67.0& 85.4& 32.1& 51.3                    & 0.32                 & 0.24                 \\
$s_1 \rightarrow s_2 \rightarrow s_4$& $\times$                                       & \textbf{68.0}& \textbf{85.7}& \textbf{33.9}& \textbf{52.1}& \textbf{0.32}                 & \textbf{0.20}\\
$s_1 \rightarrow s_2 \rightarrow s_3 \rightarrow s_4$& $\times$                                       & \textbf{68.0}& 85.6& 33.4& 52.0& \textbf{0.32}                 & 0.23                 \\ \hline
$s_1 \rightarrow s_4$& $\checkmark$                                   & 52.5                      & 77.2& 27.9& 43.1& 0.36                 & 0.28                 \\
$s_1 \rightarrow s_2 \rightarrow s_4$& $\checkmark$                                   & 55.8                      & 78.7                    & \textbf{29.8}& \textbf{46.6}& \textbf{0.34}                 & \textbf{0.27}                 \\
$s_1 \rightarrow s_2 \rightarrow s_3 \rightarrow s_4$& $\checkmark$                                   & \textbf{55.9}                      & \textbf{79.5}& 29.6& 45.8& \textbf{0.34}                 & 0.28    \\       
\bottomrule
\end{tabular}
}
\end{table}

\section{Additional Discussion}
\label{sec:addition-discussion}
\noindent \textbf{Differences from 2D Approaches.} 
While inspired by coarse-to-fine approaches in 2D image processing\cite{ho2020denoising, nichol2021improved, sohldickstein2015deep}, directly applying them to 3D presents significant challenges due to the added dimension, resulting in more complex data and increased computational demands. Furthermore, we focus on generating large-scale outdoor 3D scenes rather than the more prevalent generation of individual 3D objects. Outdoor scenes imply a higher level of diversity, comprising numerous objects, all of which require semantic coherence within the generated environments.

Another consideration is the relative scarcity of high-quality 3D datasets compared to the more mature field of 2D images. These constraints pose challenges for diffusion models in scene generation. To address this, we adopt a multi-scale approach. Initially, diffusion models train efficiently on small-scale data, ensuring diverse and semantically accurate scene generation. We then employ conditional generation techniques to refine the scenes progressively. Diffusion models excel under conditions' guidance, allowing for high-quality scene generation.

The flexibility offered by our pyramid approach ensures the diversity and quality of the generated scenes and facilitates cross-dataset generation. Additionally, the concept of our proposed Scene Subdivision Module aids in the realization of infinite scene generation, allowing for the seamless stitching and extension of scenes beyond fixed boundaries.

In conclusion, by tailoring the diffusion process to the unique demands of 3D data and leveraging conditional inputs for refinement, our method effectively bridges the gap between 2D inspiration and 3D application, unlocking new possibilities in scene generation with efficiency and adaptability.

\noindent \textbf{Limitations.} 
Despite the notable advantages of our method in both unconditional and conditional generation compared to other methods, as well as its extension to cross-dataset and infinite scene generation, it is subject to limitations primarily stemming from the scale and collecting methods of the current outdoor 3D scene datasets~\cite{wilson2022motionsc,behley2019semantickitti,caesar2020nuscenes}. Consequently, the scenes generated by our method are constrained to the largest scale of $256 \times 256 \times 16$, although our method possesses the theoretical capability to generate larger scale. Additionally, incomplete object generation may occur in the generated scenes. Despite our efforts to mitigate this limitation through infinite scene generation, the quality of the generated results is still influenced by the characteristics of the training dataset.

